%% file: main.tex
\documentclass[10pt,twocolumn,letterpaper]{article}

\usepackage{ijcb}
\usepackage{times}
\usepackage{epsfig}
\usepackage{graphicx}
\usepackage{amsmath}
\usepackage{amssymb}
\usepackage{xcolor}
\usepackage{color}
\definecolor{mygray}{RGB}{193,203,215}
\usepackage{booktabs}
\usepackage{multirow}
\usepackage{colortbl}
\usepackage{graphicx}
\usepackage[ruled]{algorithm2e}
\usepackage{algorithmic}
\usepackage{graphicx}
\usepackage{mathtools}
\usepackage{url}
\usepackage{textcomp}
\usepackage{xcolor}
\usepackage{mathrsfs}
\usepackage{booktabs} 
\usepackage{bigstrut}
\usepackage{multirow}
\usepackage{color}
\usepackage{xcolor}
\usepackage{fancyhdr}
\usepackage{listings} 
\usepackage{changepage}
\usepackage{lipsum}
\usepackage{balance}
\usepackage{enumitem}
\usepackage{comment}
\usepackage{amssymb}
\usepackage[numbers]{natbib}

\ijcbfinalcopy 

\DeclareMathOperator*{\argmin}{arg\,min}
\ifijcbfinal\fi
\begin{document}

\title{Reliable Face Morphing Attack Detection in On-The-Fly Border Control Scenario with Variation in Image Resolution and Capture Distance}

\author{Jag Mohan Singh
\quad Raghavendra Ramachandra\\
Norwegian University of Science and Technology (NTNU), Gjøvik\\
{\tt\small E-mail: \{jag.m.singh,raghavendra.ramachandra\} @ ntnu.no
}}

\maketitle
\thispagestyle{empty}

\begin{abstract}
  Face Recognition Systems (FRS) are vulnerable to various attacks performed directly and indirectly. Among these attacks, face morphing attacks are highly potential in deceiving automatic FRS and human observers and indicate a severe security threat, especially in the border control scenario. This work presents a face morphing attack detection, especially in the On-The-Fly (OTF) Automatic Border Control (ABC) scenario. We present a novel Differential-MAD (D-MAD) algorithm based on the spherical interpolation and hierarchical fusion of deep features computed from six different pre-trained deep Convolutional Neural Networks (CNNs). Extensive experiments are carried out on the newly generated face morphing dataset (SCFace-Morph) based on the publicly available SCFace dataset by considering the real-life scenario of Automatic Border Control (ABC) gates. Experimental protocols are designed to benchmark the proposed and state-of-the-art (SOTA) D-MAD techniques for different camera resolutions and capture distances. Obtained results have indicated the superior performance of the proposed D-MAD method compared to the existing methods.
 \end{abstract}

\section{Introduction}
\begin{figure} [!ht]
\includegraphics[width=0.69\textwidth]{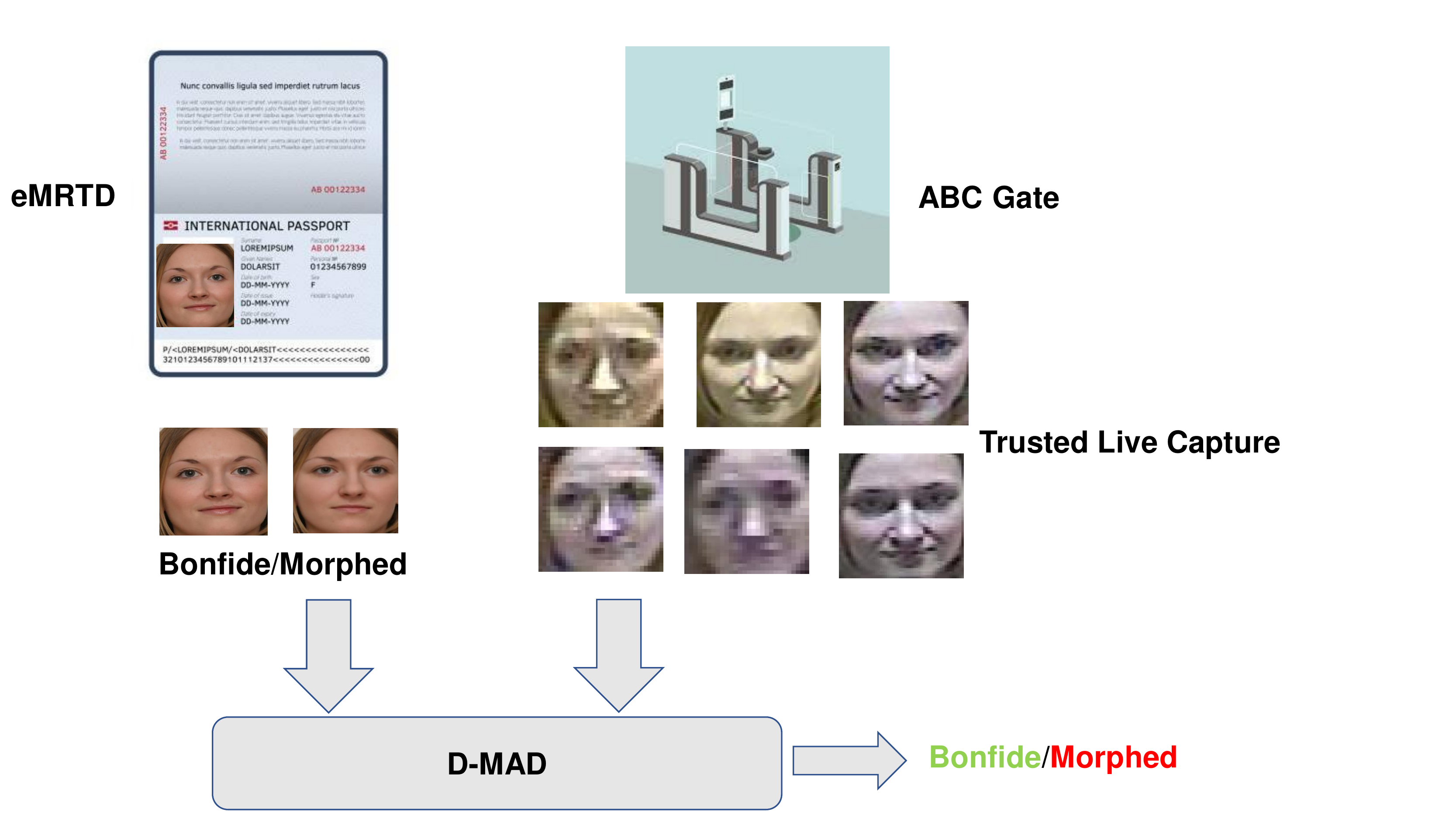}
 \caption{Illustration showing eMRTD presentation at an ABC Gate and D-MAD based decision.}
\label{fig:Intro}
\end{figure}
\begin{table*}[!ht]
\centering
\resizebox{0.98\linewidth}{!}{
\begin{tabular}{|c|c|c|c|}
\hline
{\bf{Algorithm}} & {\bf{Algorithm Classification}} & {\bf{Brief Description}}   \\ \hline
Landmark~\cite{Damer18LandmarkShiftsGCPR} & Landmark Based & Directed Landmark shifts used for classification \\ \hline
Feature-based~\cite{Scherhag18-LBPDiff-IAPR} & Feature difference based & Feature Difference used for classification \\ \hline 
Fusion of classifiers based~\cite{Damer19-FusionofClassifiers-Fusion} & Fusion of classifiers & Fusion of hand-crafted (LBPH) and CNN-based (TDCNN) features  \\ \hline 
FD-GAN~\cite{Peng19-FDGAN-IEEEAccess} & FD-GAN & Image Demorphing using symmetric dual GAN  \\ \hline
LRP~\cite{Seibold20-LRP-JISA} & LRP & Layer-Wise Relevance Propagation based on pixel-wise decision  \\ \hline
Siamese~\cite{Soleymani21-Siamese-ICPR} & Siamese Architecture & Siamese Architecture based on Inception ResNET v1 with weights from VGGFace2  \\ \hline
Mutual Information Maximization~\cite{Soleymani21-MutualInformationMaximization-WACV} & Disentanglement & Disentaglement of Appearance and Landmarks based on CNN   \\ \hline
Demorphing~\cite{Ferrara17-Demorphing-TIFS} & Image Subtraction & Inverting Morphing Equation with image-pair, and known correspondences and $\alpha$   \\ \hline
Fusion of CNN features~\cite{Singh19-Robust3DShape-SITIS} & Fusion of Classifiers & Shape (Normal-Map) and Reflectance (Diffuse Reconstruction) Decomposition: SfS-Net and Alexnet \\ \hline
DFR~\cite{Scherhag20-DFR-IEEETIFS} & DFR & Signed Distance of Arcface and Facenet features  \\ \hline
Demorphing~\cite{Ortega-Delcampo20-DemorphingCNN-IEEEAccess} & Demorphing & Autoencoder-based demorphing and face simlarity analysis  \\ \hline
Siamese~\cite{Chaudhary21_DMADWavelets_CVPR} & Siamese & Siamese for D-MAD trained on wavelet basis chosen using Kullback-Liebler Divergence (KLD)\\ \hline
Double Siamese~\cite{Borghi21_DoubleSiamese_Sensors} & Double Siamese & Double-Siamese based D-MAD, indentity-based and artifact-based  \\ \hline 
GAN~\cite{Banerjee21_CGAN_IJCB} & GAN & Conditional Identity Disentaglement using Conditional GAN for D-MAD \\ \hline
Legacy~\cite{Batskos21p_DMADLegacy_IET} & Legacy & Legacy Image and Face Verification Engine score based D-MAD \\ \hline 
\end{tabular}
}
\caption{State-of-the-art D-MAD techniques}\label{table:Algorithms}
\end{table*}
Face biometrics are widely deployed in various high-security applications, including border control, by considering usability, high accuracy, and non-intrusive capture. The high accuracy of the face biometrics can be attributed to the advances in deep-learning-based FRS methods~\cite{Schroff15_Facenet_CVPR,Parkhi15-VGG-BMVC,Deng19_Arcface_CVPR}. The exponential growth in face recognition applications has also increased the vulnerability to various attacks. Among different types of attacks on the Face Recognition Systems (FRS), the morphing attacks have mainly gained much interest due to their vulnerability in the border control scenario. The morphing process will perform the blending operation on the given face images (from contributory data subjects) to generate a single Face Morphing Image (FMI). Thus, the generated FMI includes the facial properties from all the contributory data subjects, thus demonstrating the vulnerability of both automatic FRS and human observers \cite{Rancha22_HumanObserver_Arxiv}. Since the morphing face images could be used to obtain the electronic Machine Readable Travel Document (eMRTD) or e-passports, the malicious person can exploit this process to cross the border through Automated Border Control (ABC) gates.

Face Morph Attack Detection (MAD) algorithms are extensively addressed in the biometric literature \cite{Venkatesh21_MorphingSurvey_IEETTS}. Available MAD algorithms can be classified into two main categories \cite{Venkatesh21_MorphingSurvey_IEETTS} (a) Single image based-MAD (S-MAD), where morph attacks are detected based on a single image (b) Differential MAD (D-MAD) algorithms, where morphing attacks are detected based on two or more images. Among these two approaches, the D-MAD-based MAD techniques have attracted biometric researchers by considering their application in the border control scenario. Figure \ref{fig:Intro} illustrates the D-MAD scenario in the border control application. The early work on the D-MAD approach is based on the face demorphing \cite{Ferrara17-Demorphing-TIFS}, followed by several existing methods summarised in Table~\ref{table:Algorithms}. The D-MAD approaches are developed using conventional hand-crafted features and deep features derived from pre-trained CNNs based on natural and face images \cite{Venkatesh21_MorphingSurvey_IEETTS}.

The deep learning approaches based on GAN, Siamese and Double Siamese have also been proposed for D-MAD. The benchmarking of existing D-MAD techniques is discussed in  \cite{Venkatesh21_MorphingSurvey_IEETTS, Raja20_MADDataset_TIFS} on the data captured using ABC gates indicate the severity of the problem by showing the degraded results. The ABC gate scenarios used in  \cite{Raja20_MADDataset_TIFS} are based on the one-stop such that the data subject will stand still in front of the ABC gate camera. Thus, this scenario will generate constrained images less prone to the pose and environmental (external lighting) conditions. In \cite{Singh19-Robust3DShape-SITIS}, the ABC scenario based on the 'on-the-fly' face capture 
and 3D information-based D-MAD method was introduced. Since 'on-the-fly' (OTF) face capture will result in variations in face pose, expression, and lighting, the D-MAD techniques based on face demorphing and conventional features have indicated the degraded performance \cite{Singh19-Robust3DShape-SITIS}. The experimental results with different lighting conditions indicate the further degradation of the 3D-based D-MAD results. However, it is essential to note that the existing D-MAD literature did not consider the option of different cameras with varying capture resolution and capture distance impacting the detection performance. Since the D-MAD techniques are expected to work with different ABC gates with varying camera resolutions, it is necessary to devise a suitable D-MAD method for this scenario. Thus,  we are motivated to consider the OTF ABC gate scenario with various camera resolutions and different capture distances in this work.

This work proposes a novel algorithm for a robust D-MAD, especially in the OTF ABC Gate scenario with varying image resolutions and capture distances. To this extent, we introduce a novel D-MAD algorithm based on spherical interpolation and the hierarchical fusion of deep features to detect morphing attacks. The deep features are extracted using six different pre-trained deep CNN networks that are combined using a hierarchical fusion at both the score level and feature level. Extensive experiments are carried out on the newly created database SCFace-Morph using the publicly available SCFace \cite{Grgic11_SCFace_MTA} database with 130 data subjects captured using both controlled and uncontrolled scenarios with different resolution cameras and different capture distances. We construct the new face morphing dataset  SCFace-Morph dataset using landmark-based face morphing tools ~\cite{Ferrara19-LMAUBO-Biosig} and re-digitize (or print-scan) the face morphing images to represent the real-life scenario of the border control. 

The following are the main contributions of our work: 
\begin{itemize} 
\item Proposed a novel D-MAD algorithm based on spherical interpolation and hierarchical fusion of deep features for reliable face morphing attack detection. 
\item Introduced a new face morphing dataset (SCFace-Morph) constructed using the publicly available dataset (SCFace \cite{Grgic11_SCFace_MTA}) for both digital and Print-Scan (PS) morphing attacks. 
\textit{To the best of our knowledge, this is the first work exploring morphing attack detection on the different camera resolutions and at various capture distances suitable for the OTF ABC scenario.} 
\item Extensive experiments are carried out to benchmark the performance of the proposed method with the SOTA techniques. 
\end{itemize}

The rest of the paper is organized as follows: we present the proposed method in Section~\ref{sec:Proposed}, experiments and results are discussed in Section~\ref{sec:ExpResults} and finally  conclusions and future work is discussed in Section~\ref{sec:FutureWork}.

\begin{algorithm}[!ht]
\caption{\bf{Proposed Method}}\label{scorefusionalgo} 
\hspace*{\algorithmicindent} \textbf{Input: Face Images $I_1$ and $I_2$}\\
 \SetAlgoNoLine
    \hspace*{\algorithmicindent} \textbf{Output: ($FS$ (Fused-Score))} \\
    \SetAlgoNoLine
 \SetAlgoNoLine
\begin{algorithmic}[1] 
        \STATE Compute the features from pre-trained networks for Image ($I_1$).
         \FOR{$j \gets 1$ to $6$}
          \STATE $f_1^j$ $\gets$ feature from pre-trained network.
                  \ENDFOR
        \STATE Compute the features from pre-trained networks for Image ($I_2$).
         \FOR{$j \gets 1$ to $6$}
          \STATE $f_2^j$ $\gets$ feature from pre-trained network.
           \ENDFOR
          \STATE Assign features to Groups as follows:
          \STATE $G_1$ $\gets$ $\lbrace{f_i^j}\rbrace$ where $i \in \lbrace{1,2}\rbrace$ and $j \in \lbrace{1\ldots3}\rbrace$ 
          \STATE $G_2$ $\gets$ $\lbrace{f_i^j}\rbrace$ where $i \in \lbrace{1,2}\rbrace$ and $j \in \lbrace{4\ldots6}\rbrace$ 
         \STATE Compute the feature difference
         \FOR{$j \gets 1$ to $6$}
          \STATE $\textrm{DF}^j \gets f_1^j-f_2^j$
          \ENDFOR
          \STATE Train Linear-SVM using difference features and compute scores
          \FOR{$j \gets 1$ to $6$}
          \STATE $S_j \gets$ L-SVM($\textrm{DF}^j$)
          \ENDFOR
          \STATE Use the pre-computed pair of optimal features ($x1,y1$) and ($x1,z1$)  for $G_1$ and ($x2,y2$) and ($x2,z2$) for $G_2$. They are computed once using Equation~\ref{Eqn:OP}.
          \STATE Compute SLERP (Equation~\ref{Eqn:slerpEquation}) based scores as follows where $i$ denotes Group Index and $j$ denotes the pair of optimal SLERP features inside it:
          \FOR{$i \gets 1$ to $2$}
          \FOR{$j \gets 1$ to $2$}
          \STATE $SRP_i^j \gets$ SLERP($f_i^{xj},f_i^{yj})$
          \ENDFOR
           \STATE Compute difference of SLERP features as $SRP_{D_i} \gets$ $SRP_i^1-SRP_i^2$ 
           \STATE Compute score using $S_{i+7} \gets$ L-SVM($SRP_{D_i}$)
           \ENDFOR{}
           \STATE Generate final score by fusion using sum-rule as ($FS=\sum_{j=1}^{8} S_{j}$)
           \end{algorithmic}
\end{algorithm}

\section{Proposed Method}\label{sec:Proposed}
\begin{figure*}
\begin{center}
\includegraphics[height=0.50\linewidth]{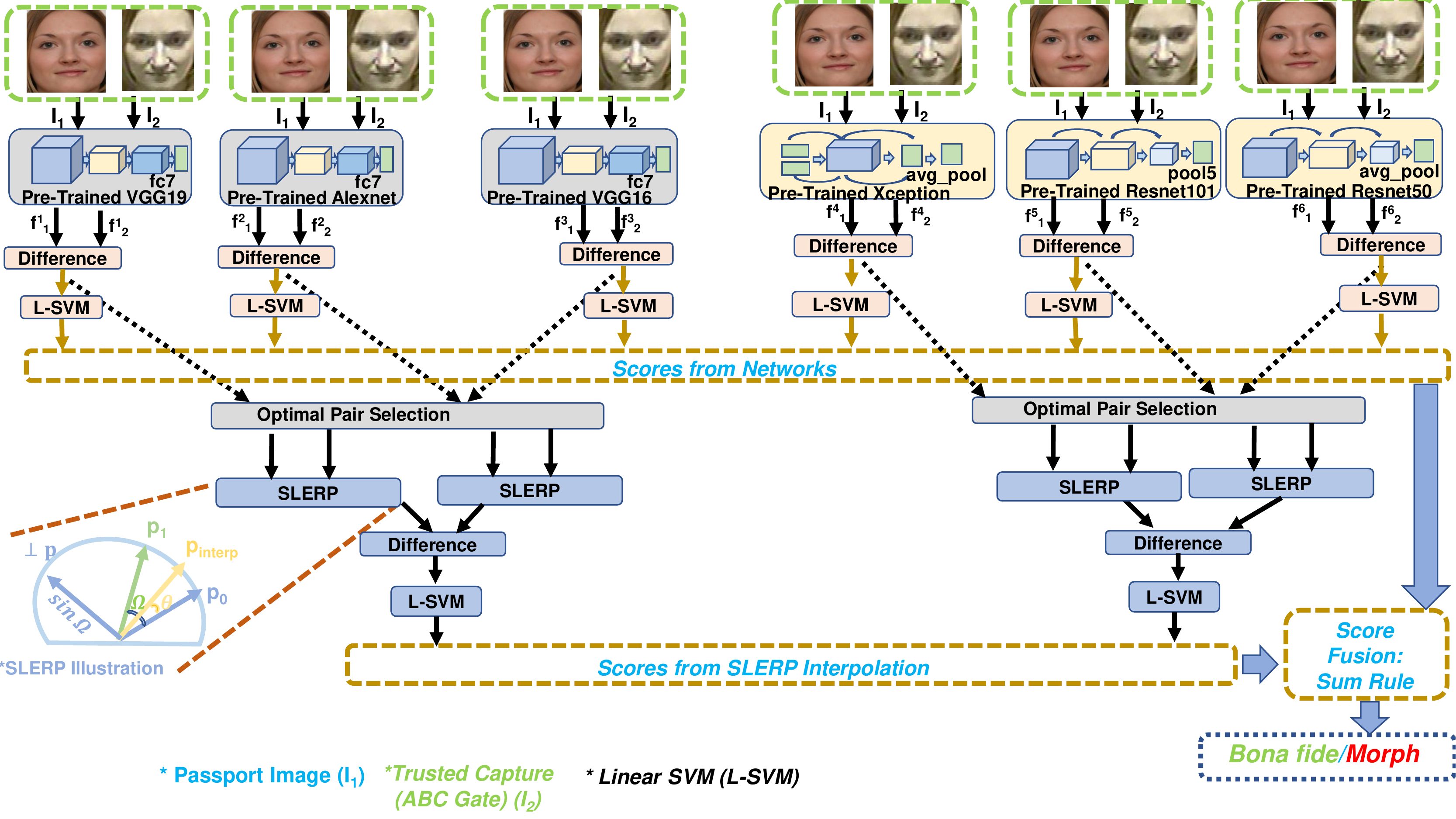}
\end{center}
   \caption{Illustration of the proposed Hybrid SLERP for Differential Morph Attack Detection (D-MAD).}
\label{fig:proposedHSLERP}
\end{figure*}

Figure \ref{fig:proposedHSLERP} shows the block diagram of the proposed method for robust D-MAD, especially in the OTF border control scenario. The proposed method is designed effectively to capture the variation of the face quality in terms of environmental changes due to lighting, pose, and expression generally encountered with the probe image by introducing a hierarchical fusion of deep features. The novel aspect of the proposed method is the feature interpolation fusion using Spherical Linear Interpolation (SLERP)~\cite{Shoemake85_SLERP_SIGGRAPH} tailored to D-MAD.   The proposed method will take two images $I_{1}$ and $I_{2}$ corresponding to the enrolment (from e-passport)  and the trusted capture (ABC Gate) face image, respectively, to detect the morphing attack on the enrolment face image. 
\begin{figure}
\begin{center}
\includegraphics[height=0.50\linewidth]{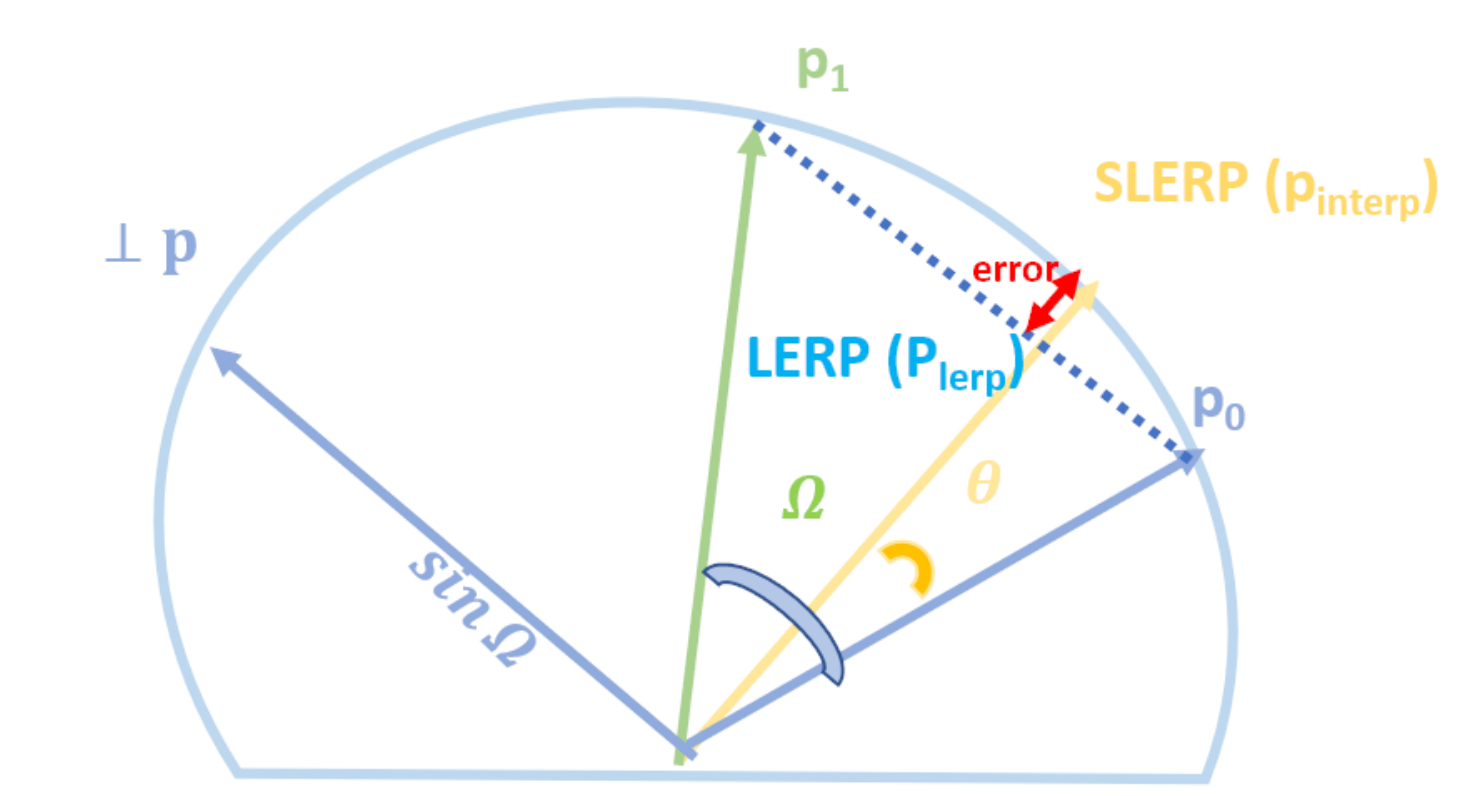}
\end{center}
   \caption{Illustration comparing SLERP and Linear Interpolation (LERP) where the error is highlighted in red.}
\label{fig:proposedHSLERPLERPComparison}
\end{figure}

The primary motivation for using SLERP~\cite{Shoemake85_SLERP_SIGGRAPH} is to perform the exact interpolation of quaternion vectors (representing a 3D rotation) on a 3D sphere. The deep features obtained for a face image are assumed to lie on a high-dimensional hypersphere \cite{Liu17_SphereFace_CVPR}. Thus, linear interpolation of deep features would lie at a point on the line joining them, but the actual interpolation should lie on the hypersphere defining the face manifold. Thus, there would be an error due to linear interpolation of deep features when compared to spherical interpolation using SLERP, which would interpolate features exactly on the hypersphere (which is also mentioned by Buss et al.~\cite{Buss01_SLERPExact_TOG} for 3D Sphere). This fact is illustrated in Figure~\ref{fig:proposedHSLERPLERPComparison} where the error is highlighted in red. Hence, we are motivated to use SLERP to interpolate deep features obtained corresponding to the face image. 
Owing to the availability of the small size face morphing datasets, the proposed method is designed using pre-trained deep CNN networks. Hence, we propose a hierarchical fusion framework to capture complementary information from different deep features effectively. The proposed D-MAD method can be structured in two functional blocks: (a) Deep feature extraction and (b) Hierarchical fusion. 
\vspace{-1mm}
\subsection{Deep feature extraction}
\label{sec:deepfeatures}
In this work, we have employed six different pre-trained deep networks trained on the ImageNet dataset~\cite{Deng09_Imagenet_CVPR}. The selected networks includes Alexnet~\cite{Krizhevsky12_Alexnet_NeurIPS}, Resnet 50 ~\cite{He16_Resnet_CVPR},  Resnet 101 ~\cite{He16_Resnet_CVPR},  Xception~\cite{Chollet17_Xception_CVPR}, and VGG16~\cite{Simonyan14_VGG_Arxiv} and VGG19~\cite{Simonyan14_VGG_Arxiv}.  These networks are selected based on their performance, and generalization for transfer learning in various applications, including morph attack detection ~\cite{Venkatesh21_MorphingSurvey_IEETTS}. Further, these six networks have indicated a good face morphing detection performance on various face morphing datasets \cite{9093488}. Since the proposed Spherical Linear Operator for feature interpolation requires the features to have identical dimensions, we need to choose the feature extraction from the pre-trained layers having identical dimensions. Hence, we made two groups of pre-trained networks where the first group ($G_{1}$) consists of VGG-19 (fc7), VGG-16 (fc7), and Alexnet (fc7) such that each network in this group will result in a feature dimension of $4096$. The second group ($G_{2}$)  consists of Xception (average pool), Resnet101 (pool5), and Resnet50 (average pool) and each of these networks will result in a feature dimension of $2048$. Thus, given the face image $I_{1}$ and $I_{2}$ from both the passport and the trusted environment (e.g., Automatic border Control Gates (ABC)), we compute the features from all six different CNN networks independently. Let the computed feature be: $F_{i} = {f_{1}^{i}, f_{2}^{i}}, \forall  i = {1, \ldots, 6}$. Where $f_{1}^{i}$ indicates the features from the passport image (or enrolment) corresponding to $i^{th}$ CNN network and $f_{2}^{i}$ corresponds to the feature from trusted  source corresponding to $i^{th}$ CNN network.
\subsection{Hierarchical fusion}
In the next step, we propose the hierarchical fusion of the features extracted from six different pre-trained deep CNN networks to achieve robust face morph detection performance. The proposed fusion scheme is implemented with both score-level and feature-level fusion of the features extracted from deep CNN networks.
The score fusion is designed with the conventional score level fusion in which the comparison scores obtained using Linear Support Vector Machines (L-SVM) based on the feature difference vector from six different CNN networks are combined using the sum rule. Given the face images $I_{1}$ and $I_{2}$, let the computed features be $f_{1}^{i}$ and $f_{2}^{i}$  $\forall  i = {1, \ldots, 6}$.  The feature difference (or residual feature computation) is performed individually for the pre-trained network $ DF^{i} = f_{1}^{i} - f_{2}^{i}$ that is then provided to L-SVM to compute the corresponding comparison score $S^{i},  \forall  i = {1, \ldots, 6}$.  
The second step is designed to perform the feature level fusion using Spherical Linear Operator (SLERP)~\cite{Shoemake85_SLERP_SIGGRAPH}. As discussed earlier in Section \ref{sec:deepfeatures}, the feature interpolation requires the identical dimension of features. Therefore, we have grouped six different networks into two main groups; in each group, we have three different networks. In this next step, we compute the optimal basis pairs for SLERP instead of all possible random combinations for each group. This way, our approach can reduce the computation and combine the complementary features. Further, the feature combination is based on feature correlation computed using residual features to achieve robustness and generalization. Thus, given the face image $I_{1}$ and $I_{2}$, we compute the residual feature corresponding to $G_{1}$ which is denoted as  $DF^{1}$, $DF^{2}$ and  $DF^{3}$. In the next step, we compute the optimal basis pair
from the triplet that can represent the complementary information to perform the residual feature combination using the SLERP method. We derive the optimal basis by computing the minimum correlation on the non-overlapping pairs generated from the given triplet, and this is indicated in Equation\ref{Eqn:OP}. 
\begin{align}
O_p = \argmin \Biggl\lbrace \rho(\textrm{DF}^{1},\textrm{DF}^{2})+\rho(\textrm{DF}^{1},\textrm{DF}^{3}),\nonumber \\
                                                \rho(\textrm{DF}^{2},\textrm{DF}^{1})+\rho(\textrm{DF}^{2},\textrm{DF}^{3}), \nonumber \\
            \rho(\textrm{DF}^{3},\textrm{DF}^{1})+\rho(\textrm{DF}^{3},\textrm{DF}^{2})) \Biggr\rbrace \label{Eqn:OP}
\end{align}
where $\rho$ indicates the correlation operation and $O_{p}$ indicates the optimal pairs of features. 
Thus, 
\begin{align}
S  = \lbrace 1,2,3 \rbrace \nonumber \\
O_{p} = \lbrace \lbrace\textrm{DF}^{\textrm{x}},\textrm{DF}^{\textrm{y}}\rbrace,\lbrace \textrm{DF}^{x},\textrm{DF}^{\textrm{z}}\rbrace\rbrace  \nonumber \\
{\textrm{s.t.}}(\textrm{x} \in \textrm{S}) \land (\textrm{y, z} \in \textrm{S}\setminus{\textrm{i}}) \land (\textrm{y} \neq \textrm{z})\label{Eqn:Optimum}
\end{align}
where, x is the index of the optimal pairs, from the triplet ${1,2,3}$ of features and y and z are the indices of the remaining features. The first optimum pair be $O_{p}^{1} = \lbrace \textrm{DF}^{x1},\textrm{DF}^{y1}\rbrace $ and second optimum pair be: $O_{p}^{2} = \lbrace \textrm{DF}^{x1},\textrm{DF}^{z1}\rbrace $. 
In the next step, we compute the SLERP feature interpolation independently for each optimum pair $O_{p}^{s}, \forall s = 1,2$ as follows:
\begin{align}
\textrm{Slerp$_{1}$}({\textrm{DF}^{\textrm{x1}}},\textrm{DF}^{\textrm{y1}},t)=\nonumber \\ \frac{\sin((1-t)\Omega)}{\sin(\Omega)}{\times}({{\textrm{DF}^{\textrm{x1}}}})) \nonumber \\
+\frac{\sin((t)\Omega)}{\sin(\Omega)}{\times}({{\textrm{DF}^{\textrm{y1}}}})
\end{align}\label{Eqn:slerpEquation}

where, $t$ is the interpolation factor which is set to $0.5$ as recommended in ~\cite{Shoemake85_SLERP_SIGGRAPH}, and $\Omega$ is the angle between the difference features ${\textrm{DF}^{x1}}$ and ${\textrm{DF}^{y1}}$ and can be computed using inverse cosine  on dot-product as $\Omega=\arccos{{\textrm{DF}^{x1}}\cdot{\textrm{DF}^{y1}}}$. { We computed the SLERP interpolation with the second optimal pair $O_{p}^{2}$ resulting in  Slerp$_{2}$($\textrm{DF}^{x1},\textrm{DF}^{z1},t$)}.  

In the next step, we perform the difference between the computed  SLERP features, which is then used to compute the comparison score $S_7$ using L-SVM for $G_{1}$. The procedures mentioned above are followed with $G_{2}$ to compute the comparison score $S_8$.
Finally, we perform the score level fusion using the sum rule to combine the scores computed from both individual CNNs and from SLERP interpolated feature differences to obtain the final score: $FS = \sum_{i = 1}^{8} S_{i} $ to make the final decision. The Algorithm of the proposed method is presented in  \ref{scorefusionalgo}.

\section{Experiments and Results}\label{sec:ExpResults}
\begin{figure*}
\begin{center}
\includegraphics[height=0.50\linewidth]{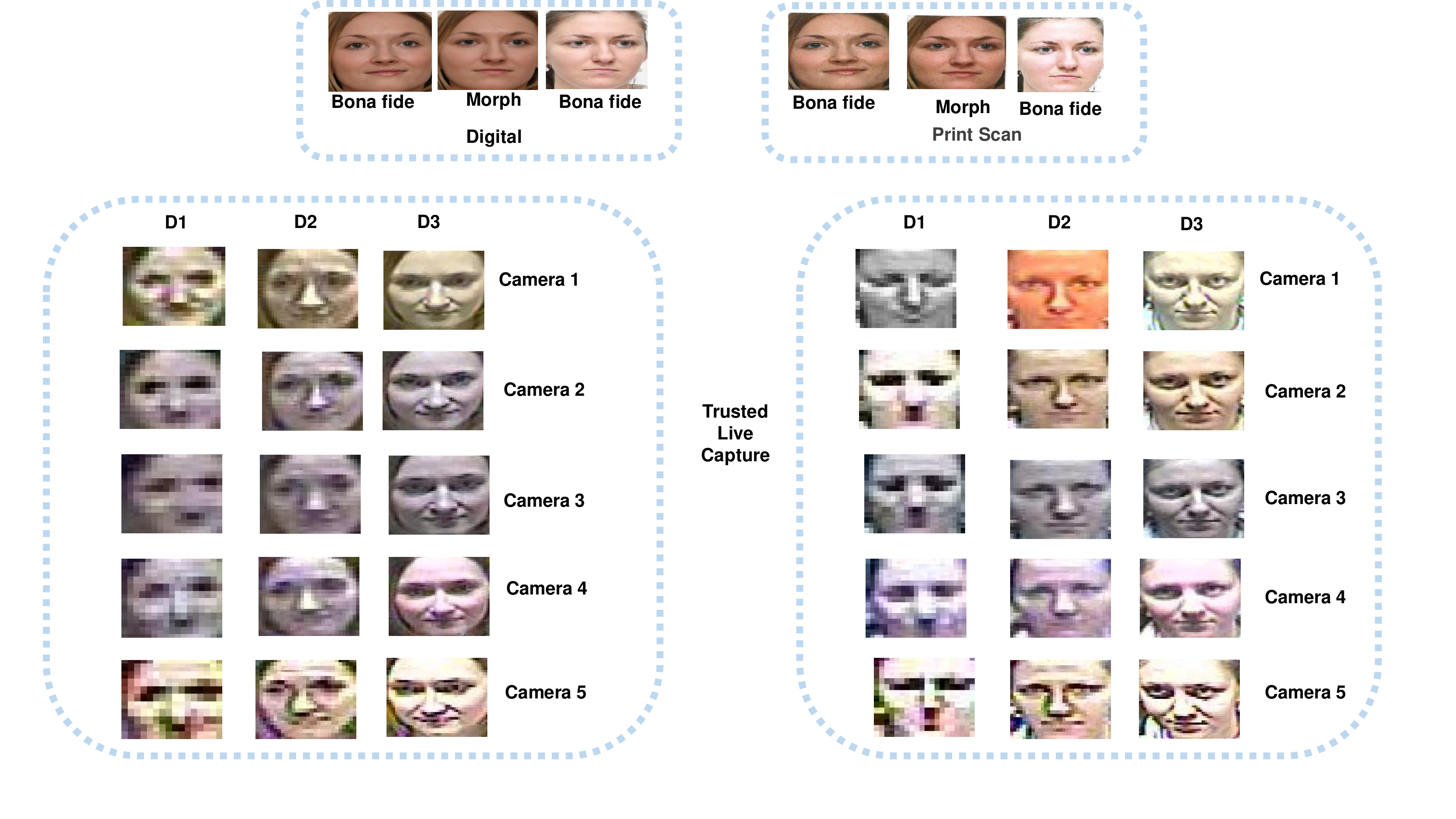}
\end{center}
   \caption{Example face images from SCFace-Morph Dataset (SCFM)}
\label{fig:datasetChallenges}
\end{figure*}
In this section, we discuss the details of the newly generated face morphing dataset based on the SCFace dataset \cite{Grgic11_SCFace_MTA}, performance evaluation protocols, and quantitative performance of the proposed D-MAD method. 

\subsection{\bf{SCFace-Morph Dataset (SCFM)}}
This work introduces a new dataset reflecting the OTF ABC systems with different camera resolutions and capture distances. We have employed the SCFace dataset by considering its applicability to real-life OTF face recognition with varying resolutions of the cameras and capture distances. The SCFace database is comprised of 130 data subjects that are captured with eight different cameras and three different distances, which are denoted as Distance-1 (D1) is 4.2m, Distance-2 (D2) is 2.6m, and Distance-3 (D3) is 1.0m. Face biometrics are captured as the data subjects walk (without a stop) through cameras held still with the frontal face capture. Since each data subject walks through these cameras, the captured data will be of unconstrained conditions with varying face poses that can represent the real-life ABC scenario. To effectively utilize the SCFace dataset for the face morphing application, we carefully selected the 77 unique data subjects by considering the image quality following ICAO standards \cite{Wolf16_PortraitMRTD_ICAO}.

Further, to reflect the real-life scenario of e-passport and ABC gates, we have considered only five different cameras that can capture visible images and high-quality mug shots. Cam1 is of resolution 540 TVL, Cam2 is of resolution 480 TVL, Cam3 is of resolution 350 TVL, Cam4 is of resolution 460 TVL, and Cam5 is of resolution 480 TVL. Thus, the mugshot images represent the face images in the e-passport. The images captured using five cameras and three different capture distances represent the trusted capture in the D-MAD algorithm evaluation. 
Next, we generate the face morphing dataset, the SCFace-Morph database (SCFM), using a mugshot image from the SCFace dataset. We have selected the neutral face pose image corresponding to each data subject to perform the face morphing using the landmark-based method from  \cite{Ferrara19-LMAUBO-Biosig} by considering its attack potential. Before performing the morphing, the dataset of 77 data subjects is divided into two independent sets, with 56 subjects in the training set and 21 subjects in the testing set. The face morphing dataset is generated following the guidelines presented in \cite{Raghavendra17_FaceAveragingvsMorphing_IJCB} that resulted in  92 face morphing images in the training set and  28 face morphing images in the testing set. The total number of probe image samples corresponding to the training set is 840 and the testing set is 315. 

The  statistics of the SCFace-Morph dataset are summarized in Table~\ref{table:tableDataset}, which specifies the number of bona fide image samples, probe image samples for each camera and capture distance and the generated face morphing images. Figure \ref{fig:datasetChallenges} shows the example face images from SCFace-Morph database dataset.  To reflect the real-life scenario of border control, we also generate the re-digitized version of both morphing and bona fide mugshot images by performing print and scan operations. We have used a Ricoh IM C6000  Color Laser multi-function printer and the scanner from the same printer. Facial images are scanned to have 300 dpi to match the requirement of ICAO standards \cite{Wolf16_PortraitMRTD_ICAO}. Thus, the newly generated database has both digital and print-scan (PS) mediums. 

\begin{table*}[!ht]
\centering
\resizebox{0.98\linewidth}{!}{
\begin{tabular}{|c|c|c|c|c|c|c|c|c|c|c|c|c|c|c|c|}
\hline
\multicolumn{16}{|c|}{\bf{Digital Images }} \\ \hline \hline
~ & \multicolumn{5}{|c|}{\bf{Bona fide Passport}} & \multicolumn{5}{|c|}{\bf{Probe Images All cameras and Distances}} & \multicolumn{5}{|c|}{\bf{Morphed Passport}} \\ \hline
{\bf{Train}}  &  \multicolumn{5}{|c|}{56} & \multicolumn{5}{|c|}{840} & \multicolumn{5}{|c|}{92} \\ \hline
{\bf{Test}}  &  \multicolumn{5}{|c|}{21} & \multicolumn{5}{|c|}{315} & \multicolumn{5}{|c|}{28} \\ \hline
\multicolumn{16}{|c|}{\bf{Probe Images per  Camera and Distance}} \\ \hline
& \multicolumn{5}{|c|}{\bf{Distance1}} & \multicolumn{5}{|c|}{\bf{Distance2}} & \multicolumn{5}{|c|}{\bf{Distance3}}  \\ \hline
   & {\bf{Cam1}} & {\bf{Cam2}} & {\bf{Cam3}} & {\bf{Cam4}} & {\bf{Cam5}}  & {\bf{Cam1}} & {\bf{Cam2}} & {\bf{Cam3}} & {\bf{Cam4}} & {\bf{Cam5}} & {\bf{Cam1}} & {\bf{Cam2}} & {\bf{Cam3}} & {\bf{Cam4}} & {\bf{Cam5}}     \\ \hline
{\bf{Train}} & 56 & 56 & 56 & 56 & 56 & 56 & 56 & 56 & 56 & 56 &  56 & 56 & 56 & 56 & 56 \\ \hline
{\bf{Test}} & 21 & 21 & 21 & 21 & 21 & 21 & 21 & 21 & 21 & 21 &  21 & 21 & 21 & 21 & 21 \\ \hline \hline
\end{tabular}
}
\caption{Statistics of SCFace-Morph Dataset (SCFM)} \label{table:tableDataset}
\end{table*}

\subsection{\bf{Performance evaluation Protocols}}
We propose three performance evaluation protocols to effectively benchmark the performance of the proposed and existing D-MAD methods by considering data medium (Digital/PS), camera resolution and capture distances. \textbf{Protocol 1} is designed to analyse the performance of the D-MAD techniques with intra and inter-medium experiments independently performed on camera and capture distance. Thus, Protocol 1 will benchmark the generalisation of the D-MAD methods for the different morph data medium and their performance impact on the image resolution and capture distance. \textbf{Protocol 2} is designed to benchmark the performance of the D-MAD techniques on intra and inter-medium irrespective of the camera resolution and the capture distance. Thus, this protocol will use all camera and distance data to train and test the D-MAD methods. Hence, this protocol will indicate the generalisation performance for a different medium. \textbf{Protocol 3} is designed to benchmark the performance of the D-MAD for individual cameras and capture distance. The D-MAD algorithms are trained and tested in this protocol by merging the Digital and PS data independently for each camera and capturing distance. This protocol will reflect the real-scenario testing as all data types are used for training the D-MAD algorithm. Hence, protocol-3 will indicate the generalisation for different camera resolutions and capture distance irrespective of data medium.

\begin{figure*}
\begin{tabular}[b]{c}
\includegraphics[width=0.22\linewidth]{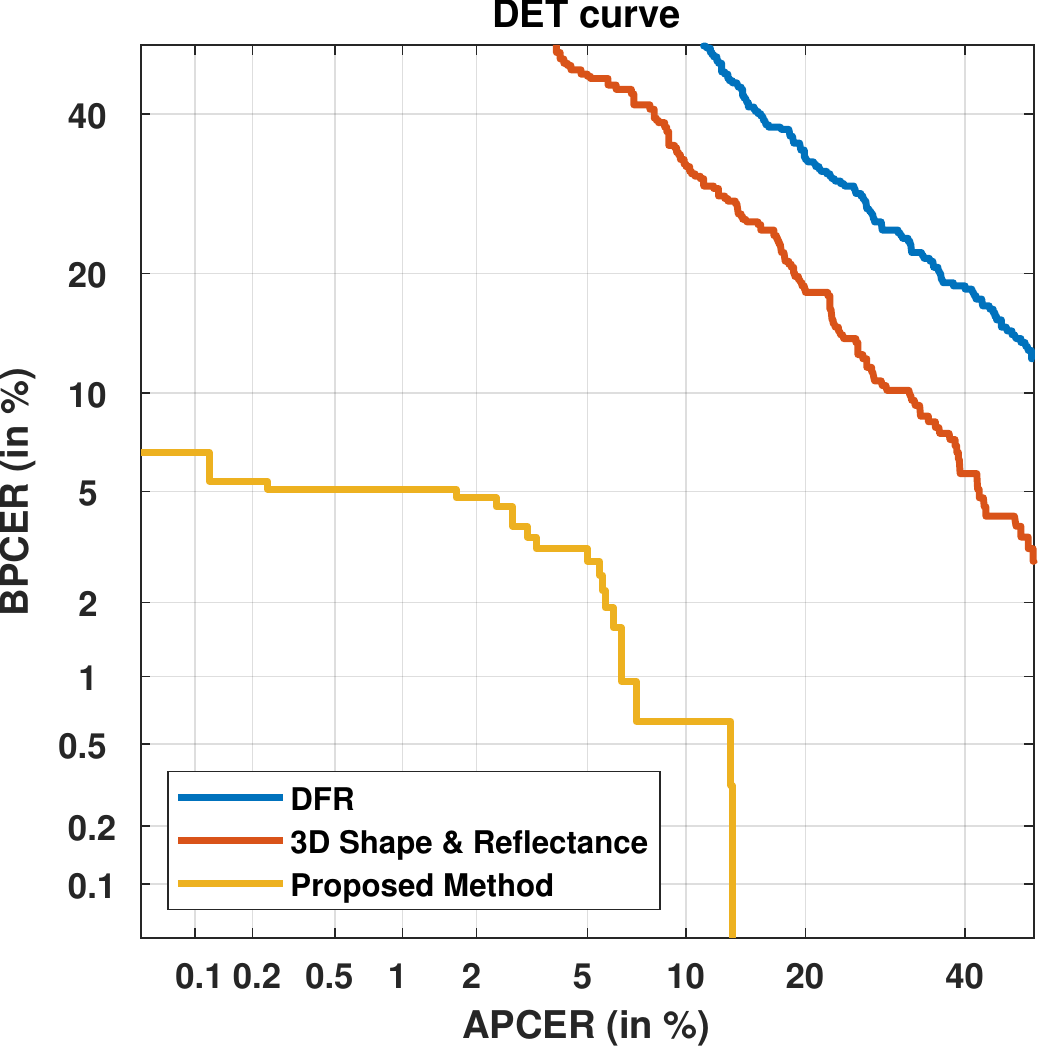} \\
\small{(a)} 
\end{tabular}
\begin{tabular}[b]{c}
\includegraphics[width=0.22\linewidth]{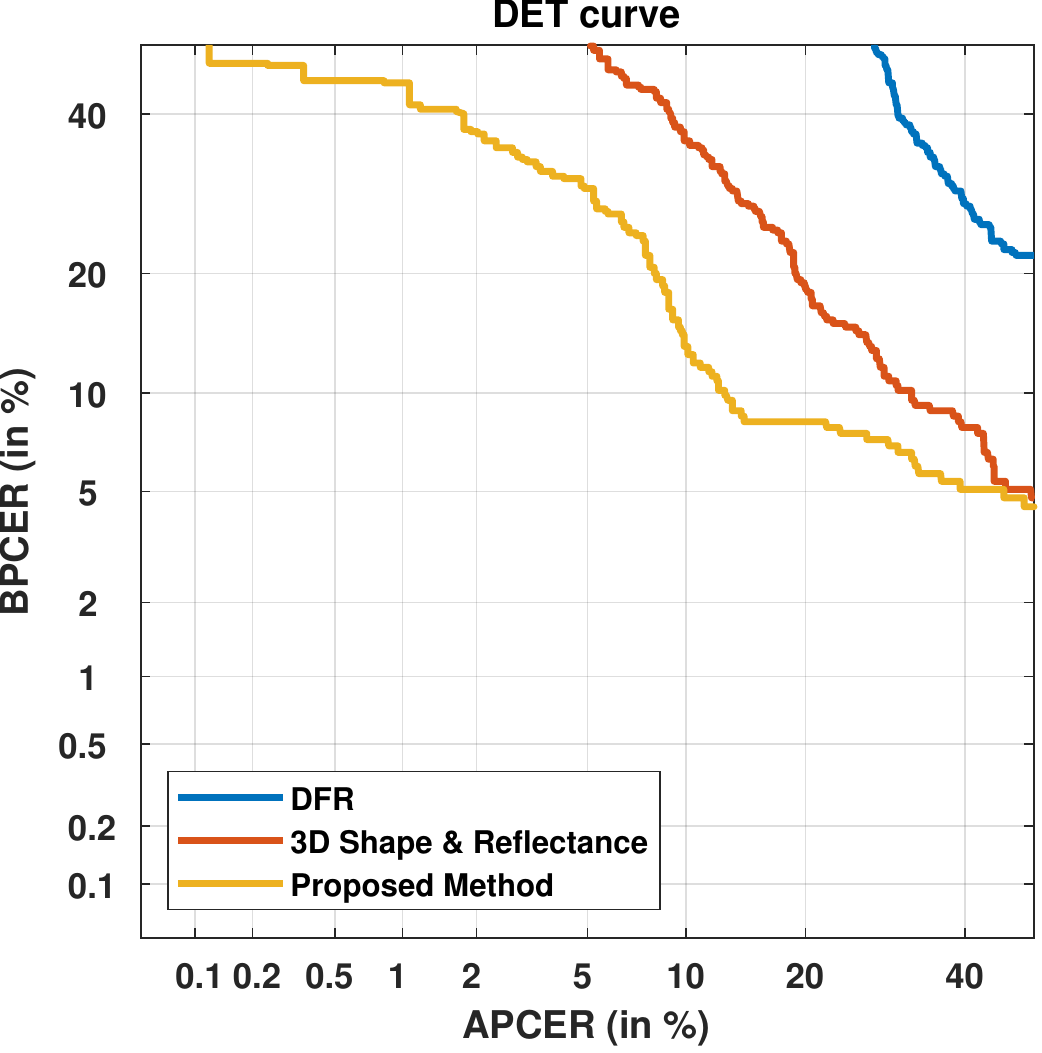} \\
\small{(b)} 
\end{tabular}
\begin{tabular}[b]{c}
\includegraphics[width=0.22\linewidth]{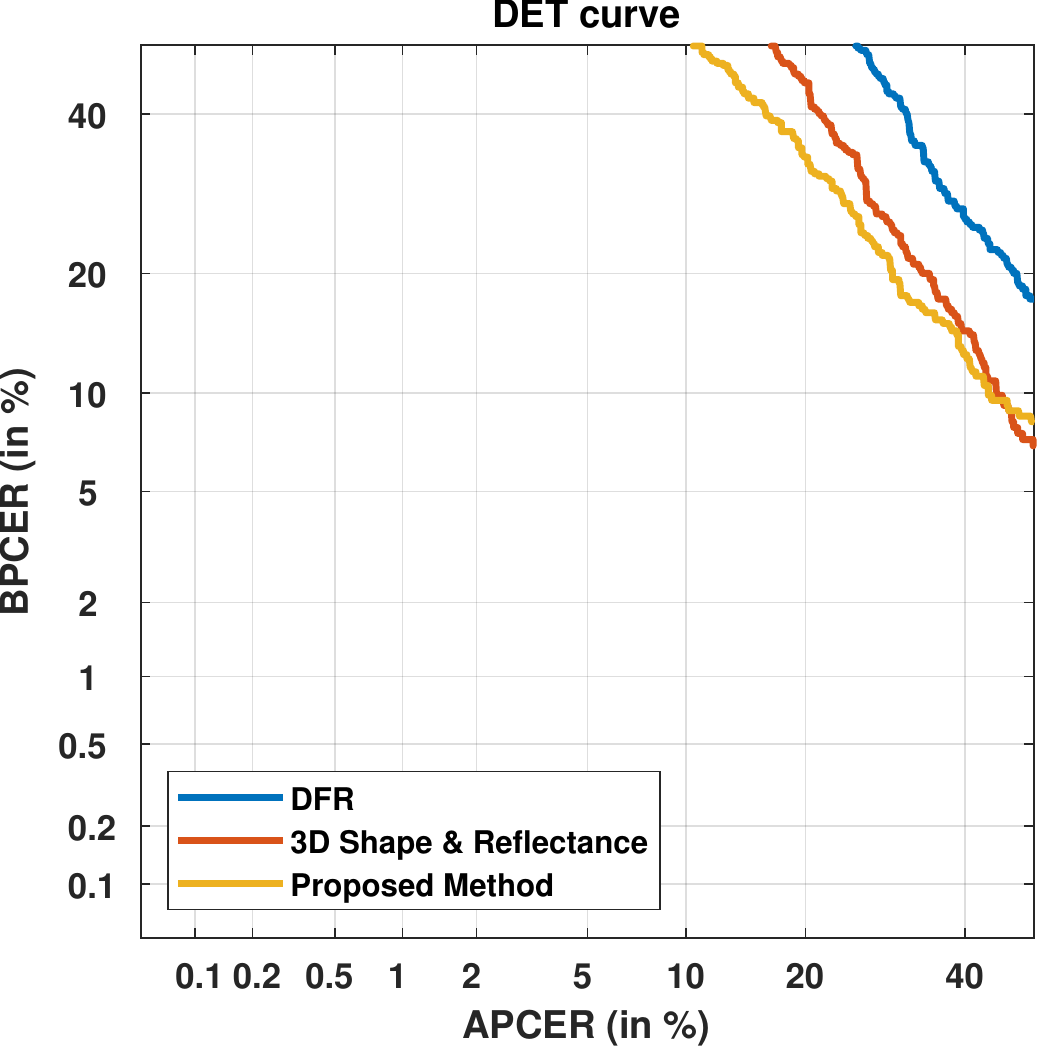} \\
\small{(c)} 
\end{tabular}
\begin{tabular}[b]{c}
\includegraphics[width=0.22\linewidth]{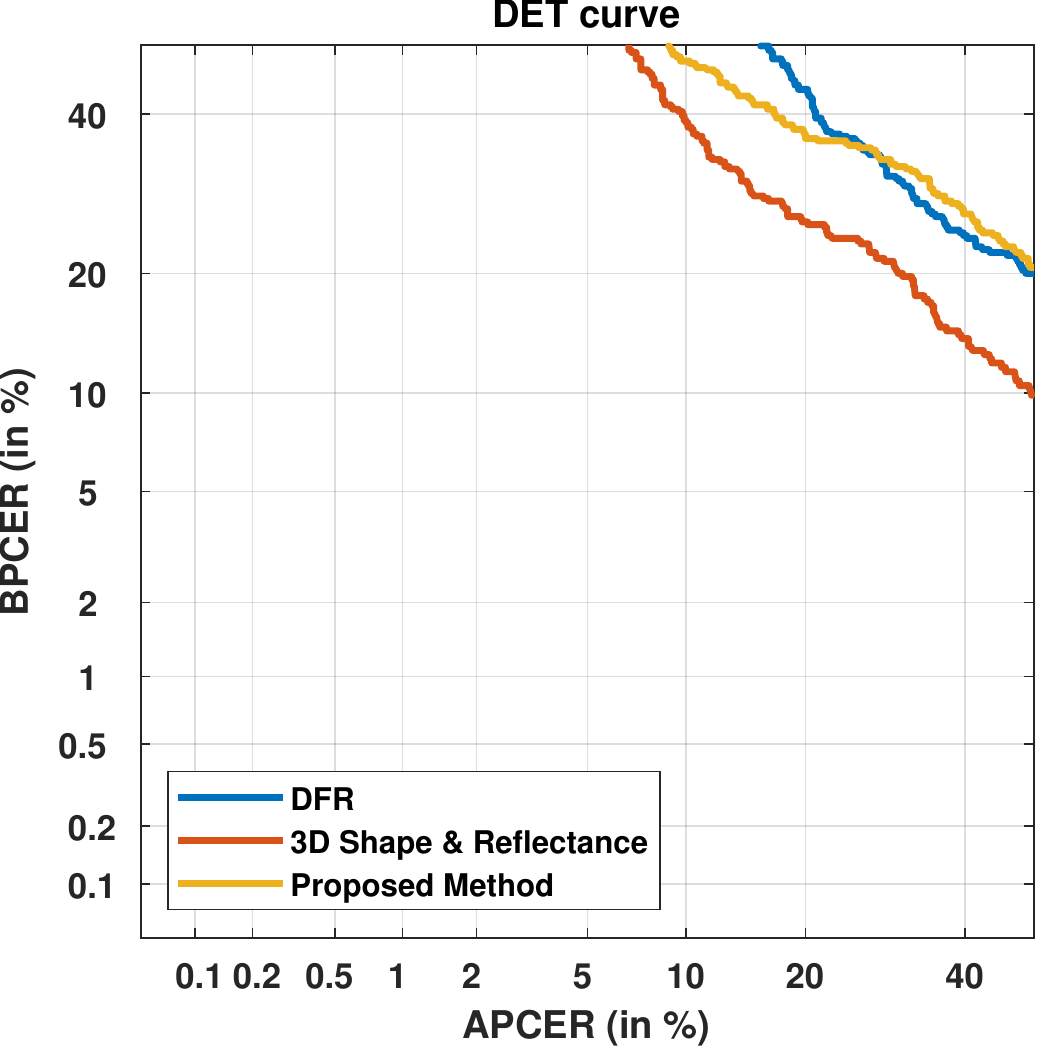} \\
\small{(d)} 
\end{tabular}
\caption{DET Curves for Protocol 2(a) Train: Digital and Test: Digital (b) Train: Print Scan and Test: Print Scan (c) Train: Digital and Test: Print Scan (d) Train: Print Scan and Test: Digital}\label{fig:DETCurves}
\end{figure*}

\begin{figure*}
\includegraphics[width=0.80\linewidth]{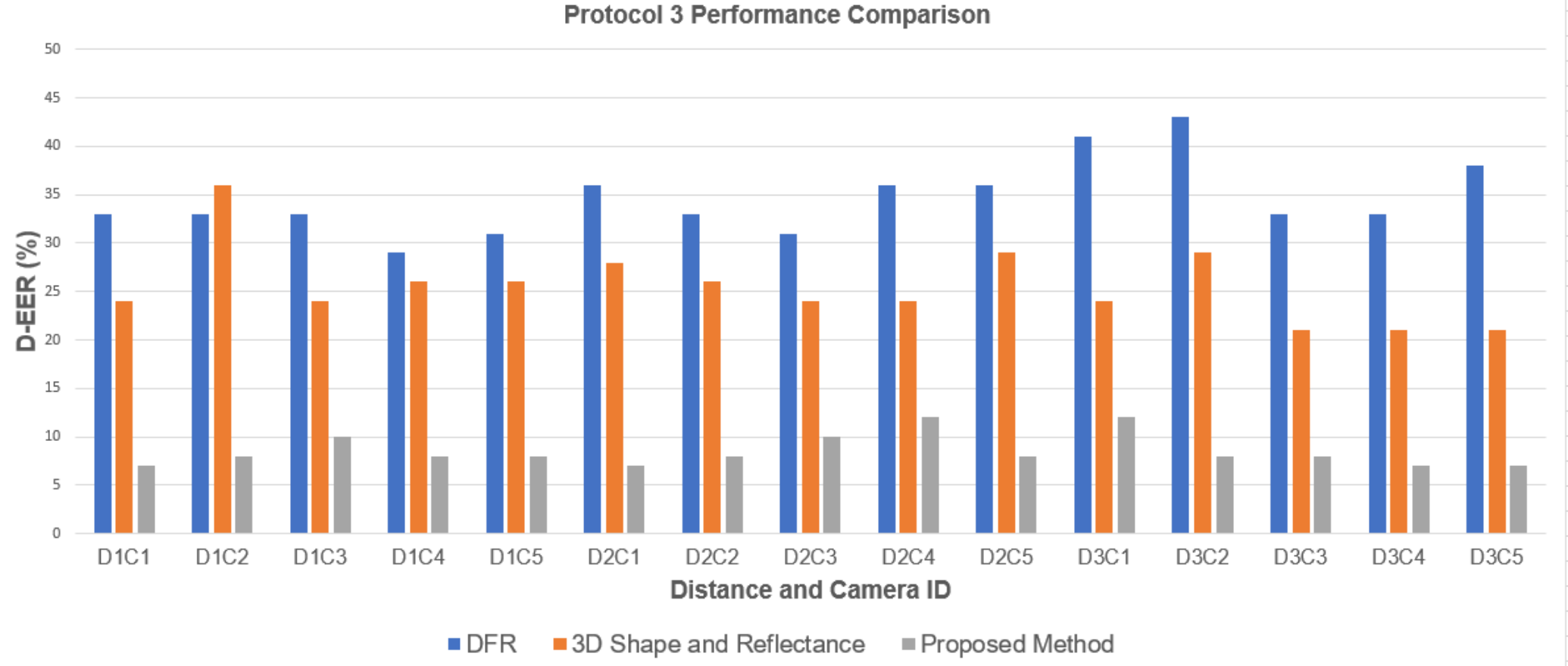}
\caption{Illustration showing D-EER for Protocol 3 for DFR~\cite{Scherhag20-DeepFaceRepresentation-TIFS}, 3D Shape and Reflectance~\cite{Singh19-Robust3DShape-SITIS}, and Proposed Method}
\label{fig:DEERProto3}
\end{figure*}

\begin{table*}[!ht]
\centering
\resizebox{0.98\linewidth}{!}{
\begin{tabular}{|c|c|c|c|c|c|c|c|c|c|c|c|c|c|c|c|}
\hline
\multicolumn{1}{|c|}{\bf{Algorithm:}}   & \multicolumn{5}{c|}{\bf{Distance1}} & \multicolumn{5}{|c|}{\bf{Distance2}} & \multicolumn{5}{|c|}{\bf{Distance3}}  \\ \cline{2-16}
\multicolumn{1}{|c|}{\bf{}}   & {\bf{Cam1}} & {\bf{Cam2}} & {\bf{Cam3}} & {\bf{Cam4}} & {\bf{Cam5}}  & {\bf{Cam1}} & {\bf{Cam2}} & {\bf{Cam3}} & {\bf{Cam4}} & {\bf{Cam5}} & {\bf{Cam1}} & {\bf{Cam2}} & {\bf{Cam3}} & {\bf{Cam4}} & {\bf{Cam5}}     \\ \cline{2-16}
& \multicolumn{15}{c|}{\bf{D-EER (\%)}}  \\ \cline{2-16}
\multicolumn{1}{|c|}{\bf{}} & \multicolumn{15}{|c|}{\bf{Train: Digital, Test: Digital}} \\ \hline  \cline{2-16}
{\bf{DFR~\cite{Scherhag20-DeepFaceRepresentation-TIFS}}} & 23.5 &23.5 &28.6 &20.2 &23.5 &28.6 &28.6 &28.6 &24.4 &28.6 &33.6 &24.4 &28.6 &28.6 &23.5 \\ \hline
{\bf{3D Shape and Reflectance~\cite{Singh19-Robust3DShape-SITIS}}} & 14.3 &23.5 &14.3 &15.2 &20.2 &19.3 &24.4 &32.7 &23.5 &19.3 &23.5 &24.4 &19.3 &23.5 &23.5 \\ \hline
{\bf{Proposed Method}} & {\bf{0.9}} &{\bf{5.1}} &{\bf{0.0}} &{\bf{8.3}} &{\bf{5.1}} &{\bf{5.2}} &{\bf{0.9}} &{\bf{5.1}} &{\bf{8.3}} &{\bf{9.2}} &{\bf{9.2}} &{\bf{4.2}} &{\bf{5.1}} &{\bf{5.1}} &{\bf{9.2}} \\ \hline
\multicolumn{1}{|c|}{\bf{}} & \multicolumn{15}{|c|}{\bf{Train: Print Scan, Test: Print Scan}}  \\ \hline  \cline{2-16}
{\bf{DFR~\cite{Scherhag20-DeepFaceRepresentation-TIFS}}} & 28.6 &29.5 &28.6 &28.6 &28.6 &42.9 &37.8 &33.6 &33.6 &27.7 &52.1 &37.8 &38.7 &42.0 &47.9 \\ \hline
{\bf{3D Shape and Reflectance~\cite{Singh19-Robust3DShape-SITIS}}} &  19.3 &23.5 &42.0 &23.5 &23.5 &19.3 &20.2 &33.6 &27.7 &23.5 &28.6 &32.7 &23.5 &28.6 &23.5 \\ \hline
{\bf{Proposed Method}} &  {\bf{10.1}} &{\bf{14.3}} &{\bf{15.2}} &{\bf{14.3}} &{\bf{9.2}} &{\bf{18.5}} &{\bf{18.5}} &{\bf{14.3}} &{\bf{19.3}} &{\bf{15.2}} &{\bf{19.3}} &{\bf{14.3}} &{\bf{13.4}} &{\bf{19.3}} &{\bf{19.3}} \\ \hline
 \multicolumn{1}{|c|}{\bf{}} & \multicolumn{15}{|c|}{\bf{Train: Digital, Test: Print Scan}}  \\ \hline  \cline{2-16}
{\bf{DFR~           } }            & 28.6&28.6 &28.6& 32.7 & 29.5&33.6&34.5&33.6&32.7&28.6&42.9&32.7&28.6 &37.8 &33.6 \\ \hline 
{\bf{3D Shape and Reflectance~}} & 23.5&37.8& 32.7&29.5 &{\bf{28.6}} &{\bf{19.3}} &38.7 & {\bf{28.6}} &28.6 &37.8 &27.7 &28.6 &24.4 &27.7 &27.7 \\ \hline
{\bf{Proposed Method}}           &  28.6      & 23.5 & 28.6 & 29.5 &29.5          &32.7     &{\bf{24.4}} & {\bf{28.6}} & 37.8 &28.6 &33.6 &       33.6 &      34.5 &37.8 &33.6 \\ \hline
\multicolumn{1}{|c|}{\bf{}} & \multicolumn{15}{|c|}{\bf{Train: Print Scan, Test: Digital}}  \\ \hline  \cline{2-16}
{\bf{DFR~\cite{Scherhag20-DeepFaceRepresentation-TIFS}}}           & 28.6       &      29.5 &28.6 &28.6 & 28.6 &47.0 &{\bf{28.6}} &{\bf{23.5}} &29.5 &{\bf{23.5}} &37.8 &{\bf{24.4}} &{\bf{28.6}} &32.7 &47.0 \\ \hline
{\bf{3D Shape and Reflectance~\cite{Singh19-Robust3DShape-SITIS}}} & {\bf{24.4}} &33.6 &42.9 &33.6 &32.7 &{\bf{27.7}} &{\bf{28.6}} &24.4 &{\bf{23.5}} &27.7 &{\bf{32.7}} &29.5 &29.5 &{\bf{28.6}} &{\bf{24.4}} \\ \hline
{\bf{Proposed Method}}                                              &  32.7      &  {\bf{23.5}} &37.8 &33.6 &33.6 &33.6 &37.8 &33.6 &37.8 &37.8 &42.9 &33.6 &33.6 &34.5 &23.5 \\ \hline
\end{tabular}
}\caption{Quantitative results of proposed method and SOTA  on  Protocol 1}\label{table:resultsProtocol1}
\end{table*}
\begin{table*}[!ht]
\centering
\resizebox{0.98\linewidth}{!}{
\begin{tabular}{|c|c|c|c|c|c|c|c|c|c|c|c|c|}
\hline
\multicolumn{1}{|c}{\bf{Algorithm:}} &  \multicolumn{3}{|c|}{\bf{Train: Digital}} & \multicolumn{3}{|c|}{\bf{Train:  Print Scan}}  & \multicolumn{3}{|c|}{\bf{Train: Digital}} & \multicolumn{3}{|c|}{\bf{Train:  Print Scan}}  \\ 
&  \multicolumn{3}{|c|}{\bf{Test: Digital}} & \multicolumn{3}{|c|}{\bf{Test:  Print Scan}}  & \multicolumn{3}{|c|}{\bf{Test:  Print Scan}} & \multicolumn{3}{|c|}{\bf{Test: Digital}} \\ \cline{2-13}
& {\bf{D-EER (\%)}} & \multicolumn{2}{|c|}{\bf{BPCER @ APCER =}} & {\bf{D-EER (\%)}} & \multicolumn{2}{|c|}{\bf{BPCER @ APCER =}} & {\bf{D-EER (\%)}} & \multicolumn{2}{|c|}{\bf{BPCER @ APCER =}} & {\bf{D-EER (\%)}} & \multicolumn{2}{|c|}{\bf{BPCER @ APCER =}} \\ \cline{3-4} \cline{6-7} \cline{9-10} \cline{12-13}
&  &   {\bf{5\%)}} & {\bf{10\%)}} &  & {\bf{5\%)}} & {\bf{10\%)}} &   & {\bf{5\%)}} & {\bf{10\%)}}  & & {\bf{5\%)}} & {\bf{10\%)}} \\ \hline   
{\bf{DFR~\cite{Scherhag20-DeepFaceRepresentation-TIFS}}}  & 27.1 & 69.5 & 56.8 & 34.7 & 93.3 & 89.2 & 34.3 & 89.2 & 84.1 & 30.9 & 84.1 & 67.3 \\ \hline
{\bf{3D Shape and Reflectance~\cite{Singh19-Robust3DShape-SITIS}}} &  19.4 & 45.4 & 32.7 &  19.4 & 50.2 & 36.2 &  27.6 & 75.2 & 60.6 & {\bf{23.8}} & 57.5 & 38.7 \\ \hline 
{\bf{Proposed Method}} & {\bf{3.4}} & 3.2 & 0.6& {\bf{11.5}} & 29.8 & 13.3& {\bf{26.0}} & 68.3 & 52.1& 32.4 & 59.4 & 47.6 \\ \hline
\end{tabular}
}\caption{Quantitative results of proposed method and SOTA  on Protocol 2} \label{table:resultsProtocol2}
\end{table*}

\begin{table*}[!ht]
\centering
\resizebox{0.98\linewidth}{!}{
\begin{tabular}{|c|c|c|c|c|c|c|c|c|c|c|c|c|c|c|c|}
\hline
{\bf{Algorithm:}}   & \multicolumn{5}{|c|}{\bf{Distance1}} & \multicolumn{5}{|c|}{\bf{Distance2}} & \multicolumn{5}{|c|}{\bf{Distance3}}  \\ \cline{2-16}
   & {\bf{Cam1}} & {\bf{Cam2}} & {\bf{Cam3}} & {\bf{Cam4}} & {\bf{Cam5}}  & {\bf{Cam1}} & {\bf{Cam2}} & {\bf{Cam3}} & {\bf{Cam4}} & {\bf{Cam5}} & {\bf{Cam1}} & {\bf{Cam2}} & {\bf{Cam3}} & {\bf{Cam4}} & {\bf{Cam5}}     \\ \cline{2-16}
\multicolumn{1}{|c|}{} &  \multicolumn{15}{|c|}{\bf{D-EER (\%)}}  \\ \cline{2-16}
\multicolumn{1}{|c|}{\bf{}} & \multicolumn{7}{c}{\bf{Train: Digital and  Print Scan}} &  \multicolumn{8}{c|}{\bf{Test: Digital and  Print Scan}} \\ \hline  \hline
{\bf{DFR~\cite{Scherhag20-DeepFaceRepresentation-TIFS}}} &32.7 &32.7 &32.7 &28.6 &30.7 &35.7 &32.7 &30.7 &35.7 &35.7 &40.8 &42.9 &32.7 &32.7 &37.8 \\ \hline 
{\bf{3D Shape and Reflectance~\cite{Singh19-Robust3DShape-SITIS}}} & 23.5 &35.7 &24.0 &26.0 &26.0 &28.1 &26.0 &24.0 &24.4 &29.0 &24.0 &28.6 &21.4 &21.0 &21.4 \\ \hline 
{\bf{Proposed Method}} & \textbf{6.7} & \textbf{7.6} & \textbf{9.7} & \textbf{7.6} & \textbf{7.6} & \textbf{7.1} & \textbf{7.6} & \textbf{9.7} & \textbf{12.2} & \textbf{7.6} & \textbf{11.8} & \textbf{7.6} & \textbf{7.6} & \textbf{7.1} & \textbf{7.1} \\ \hline
\end{tabular}
}\caption{Quantitative results of proposed method and SOTA  on Protocol 3} \label{table:resultsProtocol3}
\end{table*}

\subsection{\bf{Experimental results}}
In this section, we present the quantitative results of the proposed method together with SOTA algorithms on D-MAD, namely Deep Feature Representation (DFR) ~\cite{Scherhag19-PRNUDetection-TBIOM} and 3D Shape and Reflectance~\cite{Singh19-Robust3DShape-SITIS}. We choose the DFR method by considering its robust performance on the NIST FRVT benchmark \cite{Ngan20_FRVTMorph_NIST} and  3D Shape and Reflectance is selected by considering its application in the OTF ABC-based face morphing detection. 
The quantitative performance of the D-MAD techniques is presented using the ISO/IEC metrics \cite{ISO17_PADMetrics_ISO} namely the ``Attack Presentation Classification Error Rate (APCER ($\%$)), which defines the proportion of attack images (face morphing images) incorrectly classified as bona fide images and the Bona fide Presentation Classification Error Rate (BPCER ($\%$)) in which bona fide images incorrectly classified as attack images are counted \cite{ISO17_PADMetrics_ISO}  along with the Detection Equal Error Rate (D-EER ($\%$))´´ \cite{Zhang21_MIPGAN_TBIOM}.

Table \ref{table:resultsProtocol1} indicates the quantitative performance of the D-MAD techniques on Protocol 1. Based on the obtained results, the proposed method has indicated improved results when the medium is preserved during training and testing (Intra evaluation). Further, the proposed method has indicated the best performance in the intra-evaluation protocol irrespective of the cameras and capture distances. When the medium changes during training and testing (inter evaluation), the proposed method has indicated improved performance when training is performed on Digital and testing on Print Scan. The degraded performance of the proposed method is noted primarily in the inter-evaluation when print-scan data is used for training and digital data is used for testing. This can be attributed to the limitation of the proposed method to generalization, especially with the different image quality (because the quality of print-scan is different from that of digital) that might be due to the lack of generalized features extracted from six different pre-trained CNNs. The proposed D-MAD method generally performs better than the SOTA methods on Protocol 1.

Table \ref{table:resultsProtocol2} indicates the quantitative performance of the D-MAD techniques on Protocol 2, shown as DET curves in Figure~\ref{fig:DETCurves}. It can also be noted in this protocol that the proposed method has indicated improved performance when compared with the existing methods. The proposed method shows the best intra-evaluation protocol and comparable performance with the inter-evaluation protocol. The results indicate that the proposed method is robust to camera resolutions and capture distances.  

Table \ref{table:resultsProtocol3} shows the performance of the proposed method on Protocol 3. Based on the obtained results, it can be noted that the proposed method has indicated the best performance on both cameras and different capture distances. Further, the performance of the proposed method is not influenced by the camera type and capture distance. Figure \ref{fig:DEERProto3} graphically illustrates the  D-EER performance of the D-MAD techniques on Protocol 3.

Based on the series of experiments performed, it can be noted that the D-MAD algorithms are generally influenced by the camera resolution and the capture distance. Further, the data medium will strongly influence the performance of the D-MAD algorithms in the unconstrained ABC scenario.  
\section{Conclusions and Future-Work}\label{sec:FutureWork}
In this paper, we have presented a novel method for robust D-MAD in the ABC gate scenario. The proposed method is based on the six different pre-trained deep CNN combined using hierarchical fusion. The proposed method's novelty is in using spherical interpolation computed by SLERP to perform the residual feature fusion. Further, the hierarchical fusion is carried out using both score and feature level to achieve the robust D-MAD. Extensive experiments on the newly generated face morphing dataset (SCFM) based on the publicly available SCFace database. The performance of the proposed method and the existing techniques are extensively evaluated using three different protocols. The evaluation protocols benchmark the D-MAD performance on the different camera resolutions and the capture distance. The obtained results have demonstrated the improved performance of the proposed method in all three protocols. 
The future work includes improving the generalizability of the proposed method across different morphing image qualities especially with the different print quality. 

\bibliographystyle{unsrtnat}
\bibliography{main}

\input{supplementary}

\end{document}

%% file: supplementary.tex
\section{Supplement Material: Reliable Face Morphing Attack Detection in On-The-Fly Border Control Scenario with Variation in Image Resolution and Capture Distance}
This supplementary material presents the additional ablation results of the proposed method. We devised two experiments such that \textit{Experiment 1:}  We report both individual and intermediate results of the proposed method. \textit{Experiment 2:} This experiment is designed to indicate the efficacy of the proposed pair selection by performing the ablation study on the different pairs. In the following, we briefly discuss the outcome of the ablation study with both experiments.

\section{Quantitative results of Experiment 1}
Table \ref{table:resultsExp1Protocol1}, \ref{table:resultsExp1Protocol2} and \ref{table:resultsExp1Protocol3} indicates the performance of the proposed method and different components used to develop the proposed method evaluated in all three protocols respectively. It can be noted that: 
\begin{itemize}
    \item The performance of the individual network varies with the train and test data type. Typically, individual CNN networks perform better when trained and tested with the same data type.  
    \item Fusion of individual networks indicates the improved performance over the individual CNN networks based on the proposed pair selection algorithm. This intermediate fusion result is shown as SLERP Residue 1 and SLERP Residue 2. 
    \item The proposed method has indicated the best results compared to individual and intermediate fusion results on all three protocols. 
\end{itemize}

These quantitative results indicate the improved performance of the proposed method in all three protocols. 
\section{Quantitative results of Experiment 2}
The objective of this experiment is to justify the pair selection introduced as part of the proposed method. Since pair selection is made within the groups, we proposed the pair permutation as indicated in Table \ref{tab:pairsDesc}. Tables~\ref{table:resultsExp2Protocol1}, \ref{table:resultsExp2Protocol2} and \ref{table:resultsExp2Protocol3} indicate the quantitative results of the proposed method with different pairs in all three protocols. It can be noted that:
\begin{itemize}
    \item The proposed algorithm for the pair selection has indicated the improved performance over other possible pairs as indicated in the Table~\ref{table:resultsExp2Protocol1}, \ref{table:resultsExp2Protocol2} and \ref{table:resultsExp2Protocol3}.
    \item It can be noted that the proposed pair selection did not always show the best performance in protocol 1. However, the performance of the proposed pair is comparable. The proposed pair selection shows superior performance in average statistics, as shown in Table~\ref{table:resultsExp2Protocol1}.
    \item The proposed pair selection indicates superior performance on protocols 2 and 3. Thus, these results indicated the efficacy of the proposed pair selection algorithm, which is an integral part of the proposed method to reduce the computation without compromising the detection performance.  
\end{itemize}

Thus, experiments further justify the efficacy of the proposed method for reliable face morphing detection, especially in the mixed resolution and distance ABC scenario.

\begin{table*}[!ht]
\centering
\resizebox{0.98\linewidth}{!}{
\begin{tabular}{|c|c|c|c|c|c|c|c|c|c|c|c|c|c|c|c|}
\hline
\multicolumn{1}{|c|}{\bf{Algorithm:}}   & \multicolumn{5}{c|}{\bf{Distance1}} & \multicolumn{5}{|c|}{\bf{Distance2}} & \multicolumn{5}{|c|}{\bf{Distance3}}  \\ \cline{2-16}
\multicolumn{1}{|c|}{\bf{}}   & {\bf{Cam1}} & {\bf{Cam2}} & {\bf{Cam3}} & {\bf{Cam4}} & {\bf{Cam5}}  & {\bf{Cam1}} & {\bf{Cam2}} & {\bf{Cam3}} & {\bf{Cam4}} & {\bf{Cam5}} & {\bf{Cam1}} & {\bf{Cam2}} & {\bf{Cam3}} & {\bf{Cam4}} & {\bf{Cam5}}     \\ \cline{2-16}
& \multicolumn{15}{c|}{\bf{D-EER (\%)}}  \\ \cline{2-16}
\multicolumn{1}{|c|}{\bf{}} & \multicolumn{15}{|c|}{\bf{Train: Digital, Test: Digital}} \\ \hline  \cline{2-16}
{ Alexnet }& 14.3 & 14.3 & 14.3 & 18.5 & 19.3 & 18.5 & 14.3 & 14.3 & 14.3 & 15.2 & 19.3 & 19.3 & 15.2 & 14.3 & 10.1 \\ \hline
{ VGG16 }& 52.1 & 33.6 & 33.6 & 33.6 & 42.9 & 28.6 & 33.6 & 33.6 & 47.9 & 42.9 & 47.9 & 33.6 & 42.9 & 42.9 & 37.8 \\ \hline
{ VGG19 }& 33.6 & 33.6 & 33.6 & 23.5 & 23.5 & 29.5 & 28.6 & 42.9 & 24.4 & 33.6 & 52.1 & 47.9 & 28.6 & 29.5 & 34.5 \\ \hline
{ Resnet50 }& 14.3 & 18.5 & 15.2 & 14.3 & 14.3 & 20.2 & 19.3 & 23.5 & 23.5 & 14.3 & 9.2 & 14.3 & 9.2 & 19.3 & 23.5 \\ \hline
{ Xception }& 9.2 & 9.2 & 9.2 & 13.4 & 9.2 & 9.2 & 9.2 & 8.3 & 9.2 & 13.4 & 15.2 & 9.2 & 10.1 & 9.2 & 9.2 \\ \hline
{ Resnet101 }& 5.1 & 9.2 & 8.3 & 9.2 & 9.2 & 10.1 & 10.1 & 8.3 & 14.3 & 9.2 & 19.3 & 14.3 & 14.3 & 13.4 & 14.3 \\ \hline
{ SLERP Residue 1 }& 19.3 & 18.5 & 15.2 & 23.5 & 19.3 & 27.7 & 19.3 & 19.3 & 19.3 & 19.3 & 27.7 & 23.5 & 33.6 & 24.4 & 28.6 \\ \hline
{ SLERP Residue 2 }& 24.4 & 19.3 & 9.2 & 19.3 & 24.4 & 37.8 & 23.5 & 28.6 & 42 & 28.6 & 23.5 & 28.6 & 24.4 & 28.6 & 23.5 \\ \hline
{\bf{Proposed Method}} & {\bf{0.9}} &{\bf{5.1}} &{\bf{0.0}} &{\bf{8.3}} &{\bf{5.1}} &{\bf{5.2}} &{\bf{0.9}} &{\bf{5.1}} &{\bf{8.3}} &{\bf{9.2}} &{\bf{9.2}} &{\bf{4.2}} &{\bf{5.1}} &{\bf{5.1}} &{\bf{9.2}}\\ \hline
\multicolumn{1}{|c|}{\bf{}} & \multicolumn{15}{|c|}{\bf{Train: Print Scan, Test: Print Scan}}  \\ \hline  \cline{2-16}
{ Alexnet }& 23.5 & 15.2 & 15.2 & 23.5 & 19.3 & 24.4 & 20.2 & 28.6 & 23.5 & 23.5 & 24.4 & 23.5 & 24.4 & 20.2 & 23.5 \\ \hline
{ VGG16 }& 38.7 & 28.6 & 34.5 & 33.6 & 33.6 & 32.7 & 37.8 & 37.8 & 33.6 & 33.6 & 47.9 & 37.8 & 38.7 & 37.8 & 27.7 \\ \hline
{ VGG19 }& 42 & 28.6 & 29.5 & 23.5 & 24.4 & 23.5 & 42.9 & 28.6 & 23.5 & 37.8 & 42.9 & 33.6 & 25.3 & 42.9 & 29.5 \\ \hline
{ Resnet50 }& 28.6 & 28.6 & 42.9 & 33.6 & 28.6 & 29.5 & 37.8 & 42.9 & 33.6 & 42.9 & 33.6 & 23.5 & 33.6 & 28.6 & 33.6 \\ \hline
{ Xception }& 29.5 & 34.5 & 29.5 & 27.7 & 28.6 & 23.5 & 28.6 & 33.6 & 38.7 & 28.6 & 27.7 & 32.7 & 33.6 & 33.6 & 33.6 \\ \hline
{ Resnet101 }& 19.3 & 23.5 & 24.4 & 23.5 & 23.5 & 18.5 & 28.6 & 23.5 & 28.6 & 23.5 & 28.6 & 23.5 & 23.5 & 28.6 & 33.6 \\ \hline
{ SLERP Residue 1 }& 24.4 & 27.7 & 28.6 & 37.8 & 23.5 & 27.7 & 27.7 & 33.6 & 37.8 & 37.8 & 23.5 & 28.6 & 28.6 & 33.6 & 37.8 \\ \hline
{ SLERP Residue 2 }& 28.6 & 32.7 & 33.6 & 37.8 & 28.6 & 28.6 & 47.9 & 33.6 & 37.8 & 37.8 & 37.8 & 29.5 & 27.7 & 33.6 & 32.7 \\ \hline
{\bf{Proposed Method}} &  {\bf{10.1}} &{\bf{14.3}} &{\bf{15.2}} &{\bf{14.3}} &{\bf{9.2}} &{\bf{18.5}} &{\bf{18.5}} &{\bf{14.3}} &{\bf{19.3}} &{\bf{15.2}} &{\bf{19.3}} &{\bf{14.3}} &{\bf{13.4}} &{\bf{19.3}} &{\bf{19.3}} \\ \hline
\multicolumn{1}{|c|}{\bf{}} & \multicolumn{15}{|c|}{\bf{Train: Print Scan, Test: Digital}}  \\ \hline  \cline{2-16}
{ Alexnet }& 28.6 & 27.7 & 23.5 & 28.6 & 23.5 & 23.5 & 23.5 & 32.7 & 33.6 & 32.7 & 23.5 & 33.6 & 33.6 & 37.8 & 33.6 \\ \hline
{ VGG16 }& 38.7 & 37.8 & 28.6 & 28.6 & 37.8 & 28.6 & 37.8 & 33.6 & 32.7 & 33.6 & 38.7 & 33.6 & 37.8 & 33.6 & 37.8 \\ \hline
{ VGG19 }& 42 & 38.7 & 38.7 & 43.8 & 37.8 & 32.7 & 37.8 & 42.9 & 33.6 & 33.6 & 47.9 & 37.8 & 38.7 & 33.6 & 37.8 \\ \hline
{ Resnet50 }& 43.8 & 34.5 & 47.9 & 47.9 & 42.9 & 37.8 & 47.9 & 52.1 & 47 & 37.8 & 37.8 & 37.8 & 48.8 & 42.9 & 37.8 \\ \hline
{ Xception }& 53 & 47.9 & 52.1 & 47.9 & 52.1 & 52.1 & 57.1 & 57.1 & 47.9 & 57.1 & 47.9 & 47.9 & 52.1 & 57.1 & 48.8 \\ \hline
{ Resnet101 }& 42.9 & 52.1 & 47.9 & 47 & 47.9 & 52.1 & 43.8 & 56.3 & 47.9 & 37.8 & 58 & 52.1 & 47.9 & 58 & 47 \\ \hline
{ SLERP Residue 1 }& 28.6 & 27.7 & 38.7 & 29.5 & 27.7 & 28.6 & 37.8 & 37.8 & 42.9 & 42 & 28.6 & 42.9 & 38.7 & 34.5 & 37.8 \\ \hline
{ SLERP Residue 2 }& 47.9 & 47.9 & 56.3 & 42.9 & 52.1 & 37.8 & 38.7 & 42.9 & 57.1 & 57.1 & 38.7 & 42.9 & 52.1 & 47.9 & 48.8 \\ \hline
{\bf{Proposed Method}}           &  {\bf{28.6}}      & {\bf{23.5}} & {\bf{28.6}} & {\bf{29.5}} &{\bf{29.5}}          &{\bf{32.7}}     &{\bf{24.4}} & {\bf{28.6}} & {\bf{37.8}} &{\bf{28.6}} &{\bf{33.6}} &       {\bf{33.6}} &      {\bf{34.5}} &{\bf{37.8}} &{\bf{33.6}} \\ \hline
\multicolumn{1}{|c|}{\bf{}} & \multicolumn{15}{|c|}{\bf{Train: Print Scan, Test: Digital}}  \\ \hline  \cline{2-16}
{ Alexnet }& 37.8 & 33.6 & 28.6 & 33.6 & 33.6 & 33.6 & 28.6 & 28.6 & 33.6 & 37.8 & 33.6 & 28.6 & 37.8 & 28.6 & 28.6 \\ \hline
{ VGG16 }& 52.1 & 42.9 & 43.8 & 38.7 & 47 & 34.5 & 38.7 & 47.9 & 47.9 & 42.9 & 56.3 & 52.1 & 52.1 & 53 & 38.7 \\ \hline
{ VGG19 }& 53 & 34.5 & 47 & 39.6 & 56.3 & 42.9 & 52.1 & 42.9 & 48.8 & 57.1 & 52.1 & 56.3 & 42.9 & 53 & 33.6 \\ \hline
{ Resnet50 }& 29.5 & 33.6 & 47.9 & 29.5 & 23.5 & 28.6 & 43.8 & 42.9 & 42.9 & 42 & 34.5 & 29.5 & 37.8 & 34.5 & 33.6 \\ \hline
{ Xception }& 42 & 43.8 & 53 & 33.6 & 37.8 & 33.6 & 37.8 & 47.9 & 57.1 & 34.5 & 42.9 & 37.8 & 37.8 & 37.8 & 37.8 \\ \hline
{ Resnet101 }& 37.8 & 37.8 & 38.7 & 42.9 & 37.8 & 42 & 37.8 & 43.8 & 42.9 & 52.1 & 42.9 & 33.6 & 37.8 & 47.9 & 52.1 \\ \hline
{ SLERP Residue 1 }& 33.6 & 28.6 & 28.6 & 39.6 & 37.8 & 28.6 & 33.6 & 28.6 & 42 & 33.6 & 37.8 & 37.8 & 42.9 & 47.9 & 29.5 \\ \hline
{ SLERP Residue 2 }& 47.9 & 47.9 & 47.9 & 47.9 & 47 & 29.5 & 38.7 & 42.9 & 37.8 & 47 & 42.9 & 53 & 47 & 32.7 & 37.8 \\ \hline
{\bf{Proposed Method}}                                              &  {\bf{32.7}}      &  {\bf{23.5}} &{\bf{37.8}} &{\bf{33.6}} &{\bf{33.6}} &{\bf{33.6}} &{\bf{37.8}} &{\bf{33.6}} &{\bf{37.8}} &{\bf{37.8}} &{\bf{42.9}} &{\bf{33.6}} &{\bf{33.6}} &{\bf{34.5}} &{\bf{23.5}} \\ \hline
\end{tabular}
}\caption{Experiment1: Ablation Study Protocol 1}\label{table:resultsExp1Protocol1}
\end{table*}
\begin{table*}[!ht]
\centering
\resizebox{0.98\linewidth}{!}{
\begin{tabular}{|c|c|c|c|c|c|c|c|c|c|c|}
\hline
\multicolumn{2}{|c|}{\bf{Medium}} &  {Alexnet} & {VGG19} & {VGG16} & {Resnet50} & {Xception} & {Resnet101} & {Residue 1} & {Residue 2} & {\bf{Proposed}} \\ \hline\multicolumn{2}{|c|}{{Train: Digital,Test: Digital}} & 7.3 & 32 & 38.7 & 12 & 5.8 & 14.9 & 15.5 & 15.2 & {\bf{3.4}} \\ \hline 
 \multicolumn{2}{|c|}{{Train:  Print Scan, Test: Print Scan}} & 20.7 & 28.5 & 34.7 & 26.3 & 27 & 34.6 & 21.6 & 28.2 & {\bf{11.5}} \\ \hline 
 \multicolumn{2}{|c|}{{Train:  Digital, Test: Print Scan}} & 22.5 & 34.3 & 34.7 & 36.8 & 49.5 & 36.8 & 25.7 & 43.3 & {\bf{26.0}} \\ \hline 
 \multicolumn{2}{|c|}{{Train: Print Scan, Test: Digital}} & 40.6 & 41 & 40 & 35.2 & 37.5 & 40.3 & 31.2 & 43.8 & {\bf{32.4}} \\ \hline 
 \end{tabular}
}\caption{Experiment1: Ablation Study Protocol 2} \label{table:resultsExp1Protocol2}
\end{table*}
\begin{table*}[!ht]
\centering
\resizebox{0.98\linewidth}{!}{
\begin{tabular}{|c|c|c|c|c|c|c|c|c|c|c|c|c|c|c|c|}
\hline
{\bf{Algorithm:}}   & \multicolumn{5}{|c|}{\bf{Distance1}} & \multicolumn{5}{|c|}{\bf{Distance2}} & \multicolumn{5}{|c|}{\bf{Distance3}}  \\ \cline{2-16}
   & {\bf{Cam1}} & {\bf{Cam2}} & {\bf{Cam3}} & {\bf{Cam4}} & {\bf{Cam5}}  & {\bf{Cam1}} & {\bf{Cam2}} & {\bf{Cam3}} & {\bf{Cam4}} & {\bf{Cam5}} & {\bf{Cam1}} & {\bf{Cam2}} & {\bf{Cam3}} & {\bf{Cam4}} & {\bf{Cam5}}     \\ \cline{2-16}
\multicolumn{1}{|c|}{} &  \multicolumn{15}{|c|}{\bf{D-EER (\%)}}  \\ \cline{2-16}
\multicolumn{1}{|c|}{\bf{}} & \multicolumn{7}{c}{\bf{Train: Digital and  Print Scan}} &  \multicolumn{8}{c|}{\bf{Test: Digital and  Print Scan}} \\ \hline  \hline
{ Alexnet }& 39.3 & 39.4 & 35.9 & 39.6 & 36.5 & 38.8 & 36.3 & 40 & 40 & 39.9 & 41.1 & 37.4 & 40.2 & 39.7 & 36.9 \\ \hline
{ VGG16 }& 46.4 & 48.5 & 47.6 & 46.9 & 47.3 & 49.1 & 48.5 & 46.9 & 48.2 & 47.3 & 50.3 & 49 & 50.7 & 49.1 & 49.4 \\ \hline
{ VGG19 }& 43.6 & 42 & 41.4 & 40.5 & 37.4 & 37.8 & 40.5 & 38.1 & 40.8 & 41.7 & 46 & 45.5 & 40.3 & 42 & 37.4 \\ \hline
{ Resnet50 }& 45.5 & 42 & 46.6 & 45.5 & 38.4 & 40.3 & 43.9 & 47.9 & 46.3 & 46.3 & 40.8 & 38.5 & 43 & 37.6 & 42.9 \\ \hline
{ Xception }& 40.2 & 45.1 & 48.7 & 38.7 & 41.7 & 39.7 & 47.3 & 48.7 & 47.3 & 41.8 & 40 & 42.4 & 43.2 & 46.6 & 42.3 \\ \hline
{ Resnet101 }& 41.2 & 42 & 41.4 & 45.1 & 40.9 & 42.4 & 40.6 & 46 & 47 & 47.8 & 44.9 & 43.3 & 43.9 & 47.2 & 48.1 \\ \hline
{ SLERP Residue 1 }& 21.4 & 21.4 & 21.4 & 21.4 & 21.4 & 19.3 & 21.4 & 21.9 & 21.9 & 21.4 & 18.9 & 21 & 21.4 & 21.4 & 21.4 \\ \hline
{ SLERP Residue 2 }& 14.3 & 14.3 & 16.8 & 18.9 & 16.8 & 18.9 & 17.3 & 14.7 & 16.8 & 21.9 & 18.9 & 18.9 & 18.9 & 14.3 & 11.8 \\ \hline
{\bf{Proposed Method}} & \textbf{6.7} & \textbf{7.6} & \textbf{9.7} & \textbf{7.6} & \textbf{7.6} & \textbf{7.1} & \textbf{7.6} & \textbf{9.7} & \textbf{12.2} & \textbf{7.6} & \textbf{11.8} & \textbf{7.6} & \textbf{7.6} & \textbf{7.1} & \textbf{7.1} \\ \hline
\end{tabular}
}\caption{Experiment 1: Ablation Study Protocol 3} \label{table:resultsExp1Protocol3}
\end{table*}


\begin{table*}[!ht]
    \centering
    \resizebox{0.7\linewidth}{!}{
    \begin{tabular}{|c|c|c|}
        \hline
         {\bf{Pair Description}} & {\bf{SLERP Residue 1}} & {\bf{SLERP Residue 2}}  \\ \hline
         Proposed Method Pair & {(Alexnet,VGG16), (VGG16,VGG19)} & {(Resnet50,Resnet101),(Resnet101,Xception)} \\ \hline
         Pair 1 &   {(VGG16,VGG19), (VGG19,Alexnet) }  & {(Resnet101,Xception),(Xception,Resnet50)} \\ \hline
         Pair 2 &   {(VGG19,Alexnet),(Alexnet,VGG16) }  & {(Xception,Resnet50),(Resnet50,Resnet101)} \\ \hline
    \end{tabular}
   }\caption{Description of pairs used in the proposed method and Experiment 2}
    \label{tab:pairsDesc}
\end{table*}

\begin{table*}[!ht]
\centering
\resizebox{0.98\linewidth}{!}{
\begin{tabular}{|c|c|c|c|c|c|c|c|c|c|c|c|c|c|c|c|c|}
\hline
\multicolumn{1}{|c|}{\bf{Algorithm:}}   & \multicolumn{5}{c|}{\bf{Distance1}} & \multicolumn{5}{|c|}{\bf{Distance2}} & \multicolumn{5}{|c|}{\bf{Distance3}}  & {\bf{Mean D-EER\%}}\\ \cline{2-17}
\multicolumn{1}{|c|}{\bf{}}   & {\bf{Cam1}} & {\bf{Cam2}} & {\bf{Cam3}} & {\bf{Cam4}} & {\bf{Cam5}}  & {\bf{Cam1}} & {\bf{Cam2}} & {\bf{Cam3}} & {\bf{Cam4}} & {\bf{Cam5}} & {\bf{Cam1}} & {\bf{Cam2}} & {\bf{Cam3}} & {\bf{Cam4}} & {\bf{Cam5}}&    \\ \cline{2-17}
& \multicolumn{15}{c|}{\bf{D-EER (\%)}} &  \\ \cline{2-17}
\multicolumn{1}{|c|}{\bf{}} & \multicolumn{15}{|c|}{\bf{Train: Digital, Test: Digital}} &\\ \hline  \cline{2-17}
{\bf{Proposed Method}} & {\bf{0.9}} &{\bf{5.1}} &{\bf{0.0}} &{\bf{8.3}} &5.1 &{\bf{5.2}} &0.9 &{\bf{5.1}} &8.3 &9.2 &9.2 &{\bf{4.2}} &{\bf{5.1}} &{\bf{5.1}} &9.2 & {\bf{5.4}}\\ \hline
{Pair 1} & 5.1 & 9.2 & 0.9 & 5.1 & 8.3 & 14.3 & 0 & 5.1 & 0 & 5.1 & {\bf{8.3}} & 5.1 & 9.2 & 8.3 & 9.2 & 6.2\\ \hline
{Pair 2}& 5.1 & 9.2 & 4.2 & 9.2 & {\bf{4.2}} & 9.2 & {\bf{0}} & 5.1 & {\bf{0}} & {\bf{5.1}} & 9.2 & 4.2 & 5.1 & 5.1 & {\bf{8.3}} & 5.5\\ \hline
\multicolumn{1}{|c|}{\bf{}} & \multicolumn{15}{|c|}{\bf{Train: Print Scan, Test: Print Scan}} &  \\ \hline  \cline{2-16}
{\bf{Proposed Method}} &  {\bf{10.1}} &{\bf{14.3}} &{\bf{15.2}} &{\bf{14.3}} &{\bf{9.2}} &18.5 &{\bf{18.5}} &{\bf{14.3}} &{\bf{19.3}} &15.2 &19.3 &14.3 &{\bf{13.4}} &{\bf{19.3}} &19.3 & {\bf{15.6}}\\ \hline
{Pair 1}& 15.2 & 14.3 & 23.5 & 19.3 & 9.2 & 14.3 & 20.2 & 19.3 & 27.7 & 15.2 & 20.2 & {\bf{10.1}} & 19.3 & 19.3 & 19.3 & 17.8\\ \hline
{Pair 2}& 13.4 & 15.2 & 23.5 & 19.3 & 13.4 & {\bf{13.4}} & 19.3 & 19.3 & 23.5 & {\bf{14.3}} & {\bf{18.5}} & 15.2 & 15.2 & 23.5 & {\bf{15.2}} & 17.5\\ \hline

\multicolumn{1}{|c|}{\bf{}} & \multicolumn{15}{|c|}{\bf{Train: Digital, Test: Print Scan}}  &\\ \hline  \cline{2-16}
{\bf{Proposed Method}}           &  28.6      & {\bf{23.5}} & {\bf{28.6}} & 29.5 &29.5          &32.7     &{\bf{24.4}} & {\bf{28.6}} & 37.8 &28.6 &{\bf{33.6}} &       {\bf{33.6}} &      {\bf{34.5}} &{\bf{37.8}} &{\bf{33.6}} & {\bf{31.0}} \\ \hline
{Pair 1}& {\bf{23.5}} & 28.6 & 33.6 & 42.9 & 27.7 & 28.6 & 32.7 & 33.6 & {\bf{28.6}} & {\bf{24.4}} & 33.6 & 34.5 & 37.8 & 37.8 & 32.7 & 32.0 \\ \hline
{Pair 2}& 24.4 & 27.7 & 33.6 & {\bf{28.6}} & {\bf{27.7}} & {\bf{24.4}} & 29.5 & 38.7 & 34.5 & 23.5 & 37.8 & 37.8 & 38.7 & 42.9 & 33.6 & 32.2 \\ \hline
\multicolumn{1}{|c|}{\bf{}} & \multicolumn{15}{|c|}{\bf{Train: Print Scan, Test: Digital}} & \\ \hline  \cline{2-16}
{\bf{Proposed Method}}                                              &  {\bf{32.7}}      &  {\bf{23.5}} &{\bf{37.8}} &{\bf{33.6}} &33.6 &33.6 &{\bf{37.8}} &{\bf{33.6}} &{\bf{37.8}} &{\bf{37.8}} &{\bf{42.9}} &{\bf{33.6}} &{\bf{33.6}} &{\bf{34.5}} &{\bf{23.5}} & {\bf{34.0}} \\ \hline
{Pair 1}& 37.8 & 33.6 & 42.9 & 28.6 & 29.5 & 33.6 & 37.8 & 33.6 & 42.9 & 34.5 & 42.9 & 33.6 & 38.7 & 42.9 & 33.6 & 36.4\\ \hline
{Pair 2}& 33.6 & 29.5 & 42.9 & 37.8 & {\bf{23.5}} & {\bf{29.5}} & 37.8 & 37.8 & 47 & 37.8 & 47 & 33.6 & 37.8 & 42.9 & 27.7 & 36.4 \\ \hline
\end{tabular}
}\caption{Experiment2: Protocol 1 Ablation Study with Pair 1 and Pair 2 whose description is provided in Table~\ref{tab:pairsDesc}}\label{table:resultsExp2Protocol1}  
\end{table*}

\begin{table*}[htp]
\centering
\resizebox{0.5\linewidth}{!}{
\begin{tabular}{|c|c|c|c|c|}
\hline
 \multicolumn{5}{|c|}{\bf{D-EER (\%)}} \\ \hline
\multicolumn{2}{|c|}{\bf{Medium}} &  {Proposed Method} & {Pair 1} & {Pair2} \\ \hline
\multicolumn{2}{|c|}{{Train: Digital,Test: Digital}} & {\bf{3.4}} &  5 & 3.4\\ \hline 
 \multicolumn{2}{|c|}{{Train:  Print Scan, Test: Print Scan}} & {\bf{11.5}} &  14.6 & 13.4 \\ \hline 
 \multicolumn{2}{|c|}{{Train:  Digital, Test: Print Scan}} & {\bf{26.0}} &  28.5 & 28.2 \ \\ \hline 
 \multicolumn{2}{|c|}{{Train: Print Scan, Test: Digital}} & {\bf{32.4}} &  37.8 & 33.7\\ \hline 
 \end{tabular}
}\caption{Experiment2: Protocol 2 Ablation Study with Pair 1 and Pair 2 whose description is provided in Table~\ref{tab:pairsDesc} } \label{table:resultsExp2Protocol2}
\end{table*}

\begin{table*}[htp]
\centering
\resizebox{0.98\linewidth}{!}{
\begin{tabular}{|c|c|c|c|c|c|c|c|c|c|c|c|c|c|c|c|}
\hline
{\bf{Algorithm:}}   & \multicolumn{5}{|c|}{\bf{Distance1}} & \multicolumn{5}{|c|}{\bf{Distance2}} & \multicolumn{5}{|c|}{\bf{Distance3}}  \\ \cline{2-16}
   & {\bf{Cam1}} & {\bf{Cam2}} & {\bf{Cam3}} & {\bf{Cam4}} & {\bf{Cam5}}  & {\bf{Cam1}} & {\bf{Cam2}} & {\bf{Cam3}} & {\bf{Cam4}} & {\bf{Cam5}} & {\bf{Cam1}} & {\bf{Cam2}} & {\bf{Cam3}} & {\bf{Cam4}} & {\bf{Cam5}}     \\ \cline{2-16}
\multicolumn{1}{|c|}{} &  \multicolumn{15}{|c|}{\bf{D-EER (\%)}}  \\ \cline{2-16}
\multicolumn{1}{|c|}{\bf{}} & \multicolumn{7}{c}{\bf{Train: Digital and  Print Scan}} &  \multicolumn{8}{c|}{\bf{Test: Digital and  Print Scan}} \\ \hline  \hline
{\bf{Proposed Method}} & \textbf{6.7} & \textbf{7.6} & \textbf{9.7} & \textbf{7.6} & 7.6 & \textbf{7.1} & 7.6 & \textbf{9.7} & \textbf{12.2} & 7.6 & \textbf{11.8} & \textbf{7.6} & 7.6 & \textbf{7.1} & \textbf{7.1} \\ \hline
{ Pair 1}& 9.7 & 11.8 & 9.7 & 7.6 & 7.1 & 9.2 & 7.6 & 11.8 & 14.7 & 9.7 & 9.7 & 11.8 & 7.1 & 10.1 & 10.1 \\ \hline
{ Pair 2}& 9.7 & 11.8 & 11.8 & 11.8 & {\bf{7.1}} & 9.2 & {\bf{7.1}} & 10.1 & 13.8 & {\bf{7.1}} & 9.7 & 10.1 & {\bf{7.1}} & 9.7 & 10.1 \\ \hline
\end{tabular}
}\caption{Experiment2: Protocol 3 Ablation Study with Pair 1 and Pair 2 whose description is provided in Table~\ref{tab:pairsDesc}} \label{table:resultsExp2Protocol3}
\end{table*}